\documentclass[preprint,12pt]{elsarticle}

\usepackage[T1]{fontenc}
\usepackage[utf8]{inputenc}

\usepackage{lmodern} 

\usepackage{url}

\usepackage{amssymb}

\usepackage{siunitx}

\usepackage{cleveref}
\Crefname{equation}{Eq.}{Eqs.} 
\Crefname{figure}{Fig.}{Figs.}

\usepackage{todonotes}

\journal{Acta Astronautica}

\begin{document}

\begin{frontmatter}



\title{AI-enabled Cyber-Physical In-Orbit Factory - \\ AI approaches based on digital twin technology for robotic small satellite production}


\author[inst1]{Florian Leutert}
\author[inst1]{David Bohlig}
\author[inst1]{Florian Kempf}
\author[inst1]{Klaus Schilling}

\affiliation[inst1]{organization={Zentrum für Telematik e.V.},
            addressline={Magdalene-Schoch-Straße 5}, 
            city={Würzburg},
            postcode={97074}, 
            country={Germany}}

\author[inst2,inst3]{Maximilian Mühlbauer}
\author[inst2,inst3]{Bengisu Ayan}
\author[inst3]{Thomas Hulin}
\author[inst3]{Freek Stulp}
\author[inst3,inst2]{Alin Albu-Schäffer}

\affiliation[inst2]{organization={School of Computation, Information and Technology, Technical University of Munich},
            addressline={Boltzmannstr.3}, 
            city={Garching b. München},
            postcode={85748}, 
            country={Germany}}
\affiliation[inst3]{organization={Institute of Robotics and Mechatronics, German Aerospace Center (DLR)},
            addressline={Oberpfaffenhofen}, 
            city={Wessling},
            postcode={82234}, 
            country={Germany}}
            
\author[inst4]{Vladimir Kutscher}
\author[inst4]{Christian Plesker}
\author[inst4]{Thomas Dasbach}
\author[inst4]{Stephan Damm}
\author[inst4]{Reiner Anderl}
\author[inst4]{Benjamin Schleich}

\affiliation[inst4]{organization={Product Life Cycle Management, Technical University of Darmstadt},
            addressline={Otto-Berndt-Str. 2}, 
            city={Darmstadt},
            postcode={64287}, 
            country={Germany}}

\begin{abstract}
With the ever increasing number of active satellites in space, the rising demand for larger formations of small satellites and the commercialization of the space industry (so-called \textit{New Space}), the realization of manufacturing processes in orbit comes closer to reality. Reducing launch costs and risks, allowing for faster on-demand deployment of individually configured satellites as well as the prospect for possible on-orbit servicing for satellites makes the idea of realizing an in-orbit factory promising. In this paper, we present a novel approach to an in-orbit factory of small satellites covering a digital process twin, AI-based fault detection, and teleoperated robot-control, which are being researched as part of the ``AI-enabled Cyber-Physical In-Orbit Factory'' project. In addition to the integration of modern automation and Industry 4.0 production approaches, the question of how artificial intelligence (AI) and learning approaches can be used to make the production process more robust, fault-tolerant and autonomous is addressed. This lays the foundation for a later realisation of satellite production in space in the form of an in-orbit factory. Central aspect is the development of a robotic AIT (Assembly, Integration and Testing) system where a small satellite could be assembled by a manipulator robot from modular subsystems. Approaches developed to improving this production process with AI include employing neural networks for optical and electrical fault detection of components. Force sensitive measuring and motion training helps to deal with uncertainties and tolerances during assembly. An AI-guided teleoperated control of the robot arm allows for human intervention while a Digital Process Twin represents process data and provides supervision during the whole production process. Approaches and results towards automated satellite production are presented in detail.

\end{abstract}




\end{frontmatter}


\section{Introduction}
\label{sec:introduction}
\subsection{Motivation for in-orbit production}
The number of active satellites in orbit around Earth is ever increasing. This is on the one hand a result of the ongoing miniaturization of satellite technologies: The small size satellite systems reduce the production efforts and launch costs, while still allowing for a formation of small satellites to achieve comparable or even better results than one large monolithic system, as were deployed in the beginning of the space era. The decreasing costs of this miniaturization process led to another effect through the increase of satellite numbers, namely the commercialisation of space operations. This so-called New Space trend saw a number of commercial competitors entering the satellite market, which has previously been dominated by government-ruled institutions. New applications and demands for satellites arose, like continuous earth observation to monitor climate change \cite{yang2013role} or establishing global low-latency communication networks~\cite{vasisht2021l2d2}, which now required unprecedented numbers of small satellite formations that have to be mass produced. Some examples of these developments include the OneWeb project of Airbus aiming to launch 900 satellites in the next few years, Amazon building up a low earth orbit telecommunication network with 3200 planned Kuiper satellites, or the Starlink project by SpaceX which involves launching 12000 satellites for its network service~\cite{Kind2020}.

One possible and logical next step for satellite production would be to manufacture these types of small satellites directly in an in-orbit factory in space. Performing production where the finished product is to be used - in space - has several advantages: 1) satellites could be deployed on-demand with drastically reduced deployment times, 2) systems would not need to be designed to withstand the vibrations and high forces during rocket launch and hence could be built of drastically lighter structures, and 3) the prohibitively high launch efforts and costs~\cite{spaceflight2022} could be avoided completely. General idea of an \textit{in-orbit factory} is to bring the required materials and modular components to space for an individualized on demand in-orbit integration. This could drastically shorten production and deployment times, and increase short-hand availability of constellations of such produced satellites. A factory in space could also prospectively be used for in-orbit servicing of existing satellites. However, the technical and logistical challenges for realizing such a space factory system are not negligible.

\subsection{Challenges for Manufacturing in Space}
The term ''manufacturing in space'' introduced by \citet{skomorohov2016orbit} is used by us implicating that at least one of the steps \textit{fabrication, assembly} or \textit{integration} occurs in space.
While some research has been performed and first approaches have been studied to realize manufacturing in space~\cite{weber2018space,kempf2021ai}, the field is not mature enough yet for a set of standards to be defined. Neither the European Cooperation for Space Standardization (ECSS) nor the Consultative Committee for Space Data Systems (CCSDS) have standards pertaining specifically to manufacturing in space. Testing procedures of satellites are often tedious and redundant, and result in long product creation cycles. Each step during manufacturing and assembly must be tested \cite{unisec2022}, documented and verified by systems suppliers, system integrators and space agencies. Reducing redundant testing while guaranteeing quality could be a contribution of the Digital Twin (\Cref{sec:digTwin}), as it stores all available information about the satellites components and allows the execution some of the tests in simulations. Furthermore, unforeseen circumstances or errors during the automated production process might require human involvement. No human will however be physically present at the In-Orbit Factory to resolve such issues in person, thus, a remote possibility of human intervention is needed. Teleoperation with various degrees of control from the human and autonomy of the robot is therefore crucial for a successful space manufacturing system~\cite{globalexp2018} (\Cref{sec:teleop}). As further technical challenge during this remote robot control, trajectory planning must account for limited workspace inside the factory, and for the dynamical effects of the robot on the factory, which makes AI-support during remote intervention helpful.

\subsection{Current research for Manufacturing in Space}
It has been shown that for various satellite components as well as for spare parts, manufacturing in space is the more cost-effective solution compared to servicing missions with replacement parts manufactured on Earth~\cite{trujillo2017feasibility}.
The recent research project MOSAR~\cite{deremetz2020mosar} aimed to design modular, reconfigurable satellites where parts can be exchanged using a re-locatable robot arm for maintenance, allowing for a longer service life time. In the PULSAR project, technologies for assembling large structures in space were studied in simulations~\cite{bissonnettesimulation} and as demonstrator~\cite{rouvinet2020pulsar}. A large mirror was assembled which allows to build much larger telescopes than those that can currently be launched given the size constraints in spacecrafts - an approach that was similarly adapted during in-space self-assembly of the James-Webb-telescope~\cite{webb2022}. The PERIOD project~\cite{estable2022period} also aimed to pave the way for such concepts to be deployed to space. It included satellite assembly, reconfiguration and verification. All of these projects, however, focused on large structures. A notable exception is the approach of the group around Uzo-Okoro~\cite{uzo2020optimization,uzo2020ground}, which aims to create a prototype factory in form of a small box for CubeSat assembly to be later deployed to ISS.

\subsection{Artificial Intelligence Methods in Space Applications}
With ever increasing sensor data obtained by recent satellites, data transfer to Earth is becoming a significant bottleneck. To tackle this challenge, neural networks have been employed in the $\phi$-sat CubeSat mission to filter out images with cloud cover, thus rendering them unusable \cite{Giuffrida_2020,murphy2022developing}. Such approaches have later been extended to also allow for an autonomous operation of CubeSats \cite{Zeleke_2023}.

Machine learning methods have also been applied to perform fault detection in satellite applications \cite{guiotto2003smart} or to find anomalies in rocket propulsion  data from a testbed \cite{schwabacher2005machine}. Intelligent methods based on parameter extraction are also used on the Mars rover Curiosity to select a target for its ChemCam instrument \cite{francis2017aegis}. For prospective assembly however, so far mainly semantic and planning methods have been employed \cite{rodriguez2021autonomous,roa2022pulsar}.

\subsection{Digital Twins in Space Applications}
The in-orbit factory in space is based on digital twin technology, which is particularly prevalent in Industry 4.0 production.  According to \cite{StarkAnderlThobenWartzack20}, a digital twin is a digital representation of a product instance and, depending on the application, the associated services. This digital representation can then be used to monitor, simulation, analyse and control the physical counterpart. The digital twin therefore has a wide range of applications and offers potential for individualised and flexible production systems in particular. 

The digital twin itself was first mentioned by NASA in 2010 and has developed considerably since then \cite{Singh2021DigitalTO}. In space applications, the digital twin has been used primarily in the Satellite System, Spacecraft Cockpit Simulator, Spacecraft Development and Maintenance, Lunar Missions and Asteroid Sample Return Missions \cite{DigitalTwinSpaceSurvey22}. In this work, the digital twin is assumed to be established and used as a basis for further applications. In particular, data provision and the bidirectional connection to the physical counterpart are the central capabilities of the digital twin \cite{SCHLEICH}.

\subsection{Previous work}
First studies on how to achieve production of small Cube satellites in space have in the past also been conducted in \cite{weber2018space}: there, methods for automation of production to create high numbers of satellites at reasonable cost were investigated. Achieving this goal required developing scalable manufacturing processes with increased use of automation and digitization, employing methods from Industry 4.0 to enable mass production while still allowing for customization of products. A digital twin was used to monitor individual satellites and their configuration during production. A first prototype of a robotic AIT (Assembly, Integration and Testing) production system was developed, where the successful application of the developed production approaches could be demonstrated, and a small satellite could be assembled by a manipulator robot from modular subsystems, showcasing a flexible, holistic approach for CubeSat assembly guided by Industry 4.0 principles.

\subsection{AI In-Orbit Factory}
 In order to improve the concept of robotic production of small satellite systems presented in this paper, an important step is to make the process more robust and adaptable to achieve the desired goal of autonomous space production. This is being researched in the ``AI-enabled Cyber-Physical In-Orbit Factory'' project. In order to achieve this, the project especially focused on employing Artificial Intelligence (AI) and learning approaches to make the production process more robust and fault-tolerant. AI technology has been identified as key to future economic developments in modern production~\cite{ki_de2020}, but has not yet been adapted to, or employed in space applications. 

Key ingredient of the AI-In-Orbit-Factory was therefore the implementation of AI methods to robustify the assembly, integration and testing steps. There, those methods represent a key enabler for autonomous planning, fault detection and mitigation, offering further options to improve the adaptive manufacturing of CubeSats. The tightly coupled and AI-enabled Digital Process Twin could enhance production by not only supervising the assembly of one satellite but also continuously adapting it with knowledge generated from the assembly of multiple satellites, thus increasing flexibility and adaptability of the whole process. As direct human intervention is not possible in a space-production scenario, an intuitive teleoperation interface was developed as a fall-back solution in case human intervention is required in the autonomous production process. This interface allows an operator on Earth to control the assembly process while being optimally assisted with intelligent virtual fixtures.
The most important novel methods developed for the AI-In-Orbit-Factory are
\begin{itemize}
\item using automated robotic production approaches for realizing AIT processes for small modular satellites for the first time,
\item the integration of adaptive process planning and AI-based optimization into the developed environment, to deal with uncertainties and make the production more robust,
\item training and testing AI inspection approaches specifically for satellite components using learned models,
\item a Digital Process Twin for monitoring, orchestrating and optimizing the production process,
\item a teleoperated robotic assembly approach assisted by adaptive virtual fixtures.
\end{itemize}
After providing an overview of the AI In-Orbit Factory concept (\Cref{sec:concept}), the developed smart AIT production process - with special focus on developed AI approaches to improve this manufacturing process - will be presented in more detail (\Cref{sec:aitProc}). We then present the teleoperation approach for human intervention (\Cref{sec:teleop}) and the Digital Process Twin (\Cref{sec:digTwin}) before summarizing our findings (\Cref{sec:summary}).


\section{AI In-Orbit Factory concept}
\label{sec:concept}
\begin{figure}
\centering
\includegraphics[width=\textwidth]{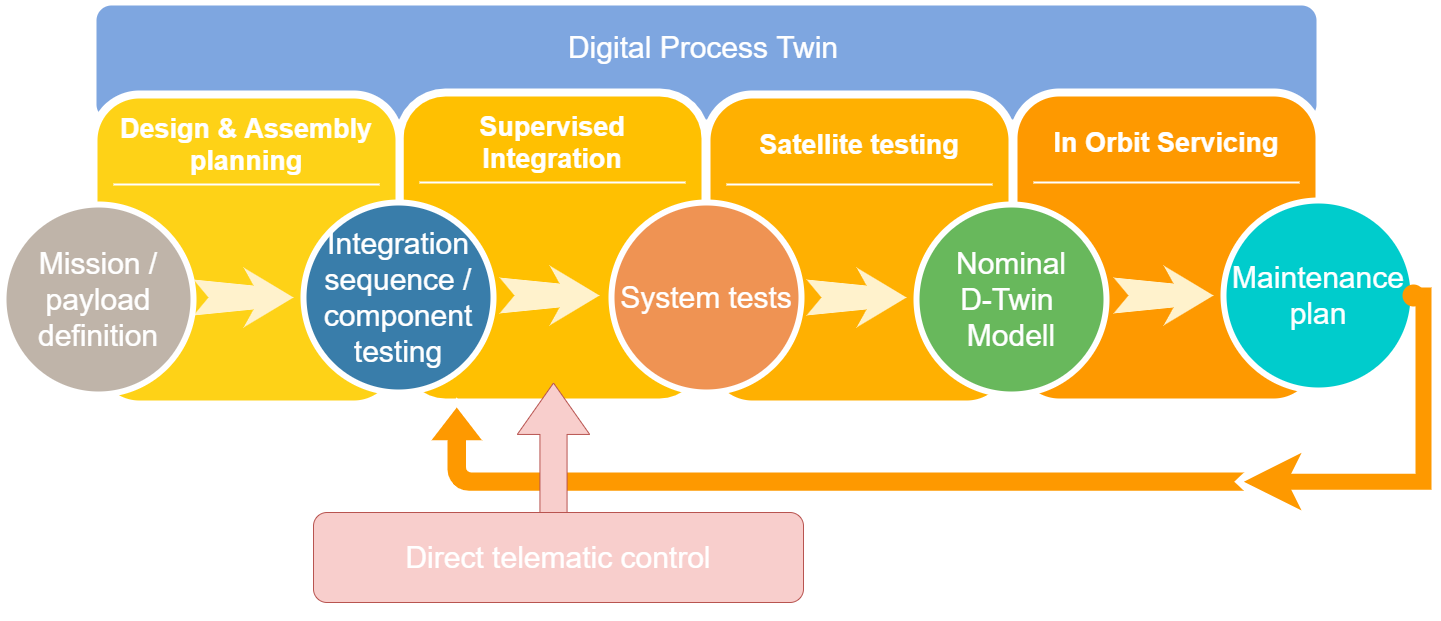}
\caption{The AIT process in the ``AI-In-Orbit-Factory'' project}
\label{fig:overview_ait}
\end{figure}

Overarching goal of the joint research project ``AI In-Orbit-Factory'' is the development of a cyber-physical production system to showcase in-orbit manufacturing, automated integration and testing (AIT) of small satellite systems, using artificial intelligence (AI)-based methods to improve adaptivity, efficiency and reliability of the production process. Central components and steps of this AIT
process can be seen in \Cref{fig:overview_ait}. 
Key algorithms for improving this AIT process include the use of Neural Networks during optical inspection, Reinforcement Learning for improving the robot's handling during integration - the use of these techniques will be detailed in the following sections.

Starting from a mission / payload definition, a central planning system first derives a component integration plan for the individual subcomponents of the satellite to be assembled. This assembly is done using a manipulator robot. During integration, every component undergoes different inspection steps, using force sensitive measuring to ensure correct handling and assembly, as well as AI-based supervision and testing methods to check for damage and ensure proper functionality. Software system tests can also be run on the fully assembled satellite before launch. 
The assembly is performed adaptively and autonomously; should human intervention be required, a teleoperation link can be established from an Earth-based ground station to directly control the robot arm. This interface can be used in case the automation fails or for assembly steps requiring the high cognitive capabilities of a human. To achieve high precision and reliability demands, AI-based shared control approaches~\cite{muehlbauer2022multiphase} leveraging probabilistic methods are used to optimally support the human operator during this demanding telemanipulation process. Finally, a passivity-based controller is used to ensure stable teleoperation despite high time delays~\cite{panzirsch2020safeinteractions,balachandran2020closing}.

All the components of this cyber-physical production system are in parallel modelled as a Digital Process Twin, providing real-time data sharing and communication between all system components as well as allowing for autonomous interaction between system parts. Furthermore, AI-based methods enable automated learning from production data to improve individual steps of the assembly process. Specifically, we implement an AI-based image recognition to detect individual satellite components during assembly as well as a state machine for flexibilisation of the assembly process.
During the process, data is being shared between all system components, from the process control PC, robot, sensors and inspection systems in the environment, as well as the Digital Process Twin and the human teleoperator in case of remote control. This is being achieved using a unified communication interface, the so-called compositor, providing real-time access to all required data between system components.
All individual components of the AI-In-Orbit-Factory production system are presented in more detail in the following, with special focus on how AI-based methods are employed to improve robustness, reliability and quality of the production process.



\section{Adaptive AI-supported AIT processes for satellite integration}
\label{sec:aitProc}
Smart manufacturing of small satellites has been a major focus in the project consortium before, with focus on realizing scalability for large scale production~\cite{schillingGrossproduktionKleinsatelliten2021a} using process permeating digitalization~\cite{kraussDigitalManufacturingSmart2021}, robotic au\-to\-ma\-tion~\cite{schillingROBOTICSEFFICIENTPRODUCTION2017,schillingAdvancedRoboticAutomation2017} and modularity with the UNISEC satellite bus. For completeness, components and processes of the core assembly system are described shortly again in \cref{sec:aitCoreComp,sec:aitCoreProc}. The additional AI approaches and support components complimenting and optimizing these AIT processes are the special focus of this contribution and thus are introduced in detail in the following paragraphs. 

\subsection{AIT production environment}
\label{sec:aitCoreComp}

\begin{figure}
\centering
\includegraphics[width=\textwidth]{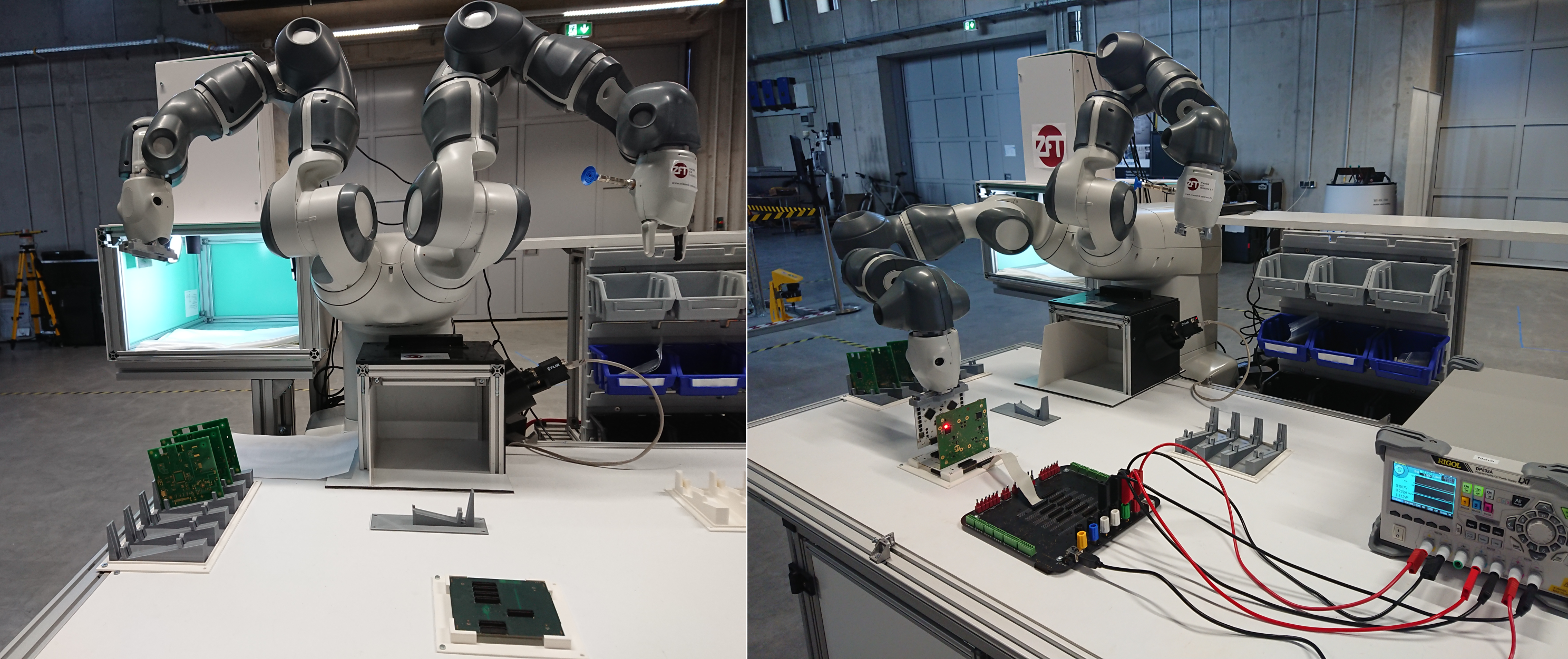}
\caption{AIT production environment with the YUMI robot for automating the integration process, optical inspection station (left) and electrical test setup (right)}
\label{fig:aitEnv}
\end{figure}

Setup and main elements of the AIT production environment can be seen in \Cref{fig:aitEnv}: Central component is the two-armed collaborative YUMI robot performing the actual automation steps. It has two 7-axis arms with integrated force-torque-sensors, allowing for good reachability and force-sensitive handling of all components inside the work area. It has been equipped with specially designed grippers ensuring safe and precise handling of the 1U-satellite subsystem boards used in the demonstration system. Several tray and handling holders for satellite components are shown in \Cref{fig:aitEnv} as well: although rigidly mounted relative to the robot, tolerances in the components' placement can also be resolved and handled autonomously during assembly. Also shown is the optical inspection station consisting of 4 cameras (one overview and macro industrial camera as well as two electronic microscopes providing different object resolutions) in a shielded environment using controlled lighting conditions as well as a polarized crosspole filter to avoid reflections. The cameras have been precisely calibrated relative to the robot to ensure that features on the inspected components can be localized on the board relative to the gripper / robot. The electrical testing setup can be seen in \Cref{fig:aitEnv} right, consisting of a so-called \textit{DevBoard} providing power and communication connection to subsystems mounted on the satellite backplane, as well as a programmable power supply unit for measuring electric values.

\subsection{Adaptive and robust production processes}
\label{sec:aitCoreProc}
Integration and testing of satellite components has been developed and tested for two application scenarios: solar cell assembly on the outer satellite panels, and integration of satellite subsystem boards into a backplane comprising the satellite communication bus. The latter part focuses more on AI approaches and thus will be described here in more detail
.

The AIT process for subsystem integration starts with placement of the backplane and a tray holding the required subsystems to be integrated into their respective holders in the production environment. In principle, the exact geometry of the robot's workspace including the holder's position relative to the robot is known. Due to this knowledge, the very limited available space in the production environment and the thus desired predictability of the robot's movements, the processing trajectories are optimally determined before, and mostly fixed during execution. However, in order to be able to robustly deal with some small deviations, positioning tolerances and uncertainties, the robot may employ certain AI-based checks and learning algorithms to slightly modify the programmed paths or the program execution. Some examples of this will be described in this section and in further detail in section \cref{sec:aitLearning}. As one example, placement of the (asymetric/non-square) subsystem boards could be subject to some geometrical deviations regarding orientation and positional error - to be as robust as possible, the AIT assembly system handles such uncertainties during assembly. This is done by using a camera integrated into the YUMI robot using optical similarity algorithms to resolve correct placement of the backplane in the assembly holder. The system can also deal with the 4 possible 90 degree-rotated insertions by adapting the insertion movements accordingly, providing maximum placement tolerance for the backplane. To safely resolve the placement of the subsystem boards, the robot positions its gripper close to the expected board side location and uses its force-sensitive grippers to determine if a tray slot contains a board at all, and in which orientation it has been inserted (boards are non-square). This is done by carefully opening the gripper until a resistance is encountered, thus measuring the distance between gripper starting location and subsystem side, and comparing it with the known dimensions of this Cubesat board. Using this information, the board is then localized and can be gripped without compromising the (yet unknown) electronic components on the board's surface, using only the structural holes (present in every Cubesat subsystem with known location relative to the board's side) as well as the board's outer sides to achieve a save and stable grip of the board. 

After a subsystem has been successfully gripped, the robot inserts it into the vision station, to determine its type and orientation as well as perform a checkup for possible pollution and defects. This AI-supported optical inspection is detailed further in \cref{sec:aitOpticalInsp}. If a system passes the inspection, the determined type is then compared with the assembly plan and inserted into the correct backplane slot, or removed from the current production cycle and put down into a separate holder tray for discarded boards. 

The actual integration of the subsystem into the backplane is performed using a force-sensitive insertion approach also supported by AI learning (details in \cref{sec:aitLearning}). Target insertion slot on the backplane is known from the assembly plan. Depending on how the subsystem was gripped initially it might need to be flipped several times in the gripper to ensure the connector on the subsystem is pointing away from the gripper, so the robot can apply the necessary force to actually insert it into the backplane. Flipping of the board in the gripper is done using the second robot arm, presenting a handling approach also feasible in Zero G. After the board has been suitably gripped, it is carefully inserted into the proper slot in the backplane with force supervision (\cref{sec:aitLearning}). Should errors occur during this assembly step, the system might request teleoperated intervention (\cref{sec:teleop}). 

After insertion, software tests as well as electrical checks using learned proper values for this board type are run to ensure proper functionality of the electronics (\cref{sec:aitElectricInsp}), using the connected DevBoard to provide the required power as well as communication with the subsystem. If a board fails during these tests, it can once again be disassembled and discarded. In these cases, the second robot arm is utilized for unplugging the board, holding the backplane down to provide the necessary stability and avoid slippage during unplugging. This ability for disassembly makes the developed AIT-system also feasible for prospective servicing of existing satellite systems.

After successful insertion and testing of the current subsystem, the system proceeds with all further boards in the loading tray, until assembly has been completed.

After this simplified overview of the AIT process, the following sections describe in more detail specific parts of this automated production as well as where AI approaches are used to make these production steps more fault-tolerant, robust and thus autonomous, or to optimize the production further.

\subsection{Neural networks for optical inspection}
\label{sec:aitOpticalInsp}
Purpose of the optical inspection during this AIT process is twofold: for once, to verify type and gripping orientation of the currently held board (stage 1). 
Second, to check the handled subsystem for pollution or damage / defects on its surface (stage 2) which might make it unsuitable for integration in the satellite system to be build.

During stage 1 of the inspection, an overview image is taken and compared to a database in the Digital Twin containing reference images of all known types of subsystems. Image processing and similarity algorithms (for example cross correlation~\cite{burger2009digitale}) are used to determine type and orientation. Identification can be done reliably due to precise positioning of the board relative to the camera by the robot, and optimal lighting conditions in the shielded setup with cross pole filtering to avoid reflections. This enables to quickly  and reliably resolve the type and orientation of the gripped board, and how to proceed with the integration.

\begin{figure}
\centering
\includegraphics[width=\textwidth]{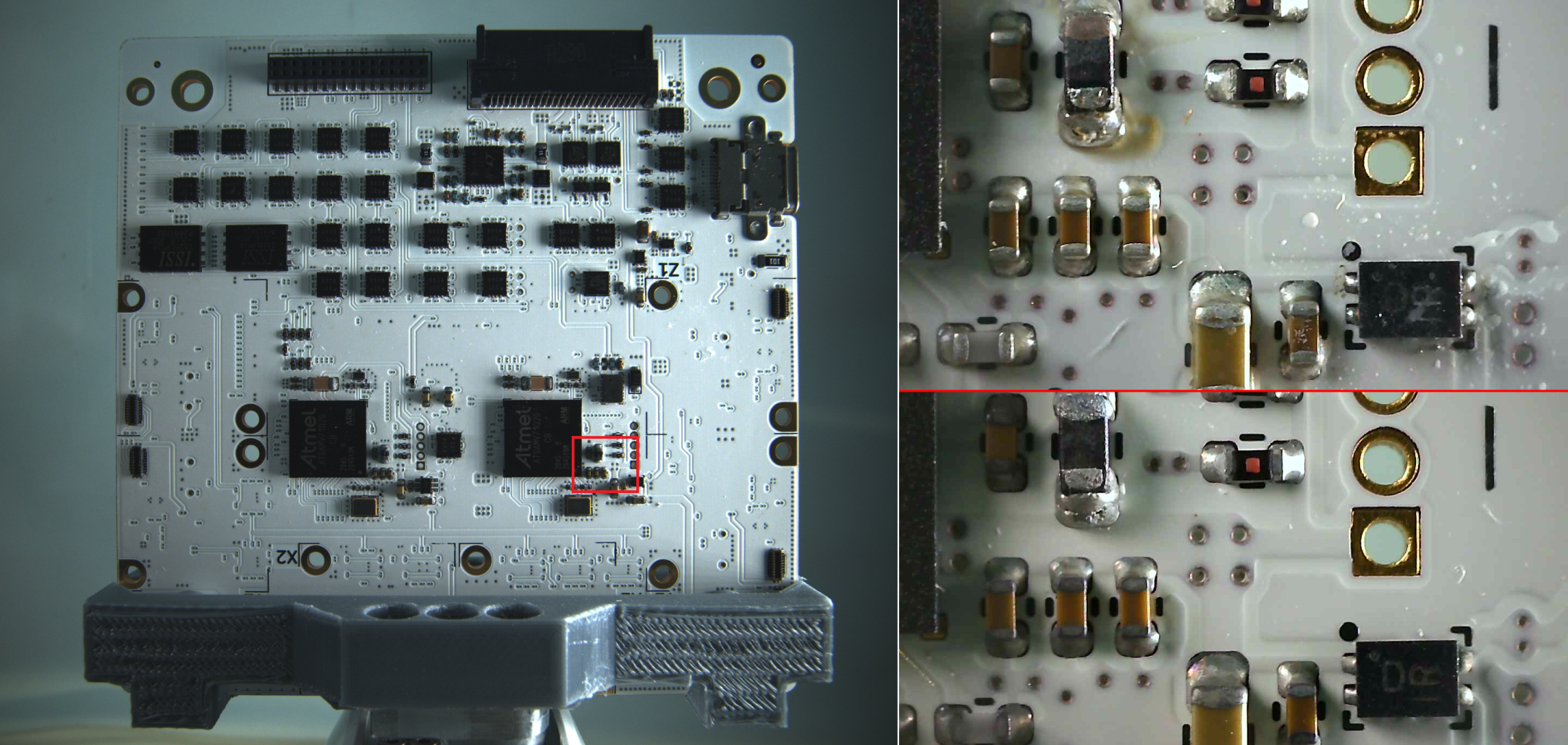}
\caption{Overview image used to identify board type and orientation during optical inspection (left); microscope view of the indicated area (right), of a polluted board (top) and after cleaning (bottom).}
\label{fig:optInsp0}
\end{figure}

In parallel, the overview image of the board is also inspected for any deviations from its expected appearance (see \Cref{fig:optInsp0}). Thresholding limits in the similarity checks determine possible defects or pollution, in which case stage 2 of the optical inspection is started for the determined regions of interest. 

During stage 2, detailed macro or microscope images are taken from suspect areas on the board, once again using the robot to precisely position the board under those cameras accordingly. These detailed inspection images are then analyzed by an AI system to identify visible features on the board. A semantic segmentation network trained on a data set of 744 images created by the authors by running inspection on different types of subsystems is used to create an initial set of segmentation masks for the five chosen categories: components, solderpads, solderbridges, solderballs and tombstones. After the initial prediction the probability maps are binarized and the contours of individual object instances are extracted using a contour following algorithm~\cite{SUZUKI198532}. 
The input images 
are fed through the segmentation network to predict the class memberships of each pixel. It is also possible for each pixel to belong to multiple of the five classes as solder-covered components are a common occurrence and some solder balls might be on top of components or solder joints. After thresholding by a class-specific value the contours of all objects are extracted from the binary masks. Finally the detected objects and their bounding boxes are returned. 

%

The segmentation network is based on a modified UPerNet~\cite{upernet2021} with ConvNeXt layers~\cite{convnext} instead of default convolutional layers. It was trained for 1000 epochs on 16.100 annotated components of satellite PCBs with strong data augmentation, comparable to those of imaug~\cite{imgaug} and batch size 4 on a RTX 2070 GPU with a soft-focal-loss~\cite{SoftFL}. 

The 
trained model can identify most classes with high precision and recall values, with only the ``Solderball'' class performing less than optimal: The chosen soft-focal-loss results in a strong weight for this class during training, as positive samples often contain $>99\%$ background pixels. This results in a network that is highly sensitive to solderballs and already generates a lot of false-positive samples purely to minimize the training loss. Additionally the solderball class is the least precisely labeled class in the dataset, with annotators sometimes overlooking samples or labeling dust particles or solder-flux residue. As dust and grainy surfaces are often falsely identified as solderballs by the labeling workers the network will also be sensitive to these features, increasing the false-positive rate. 
Some labels contain a solderball and a considerable amount of background, increasing false-negative classifications. Despite these small limitations, the trained  Neural Network showed significant success in detecting possible defects and pollution even on previously unseen satellite components and was thus integrated into the AIT process to improve its reliability.

\begin{figure}
\centering
\includegraphics[width=\textwidth]{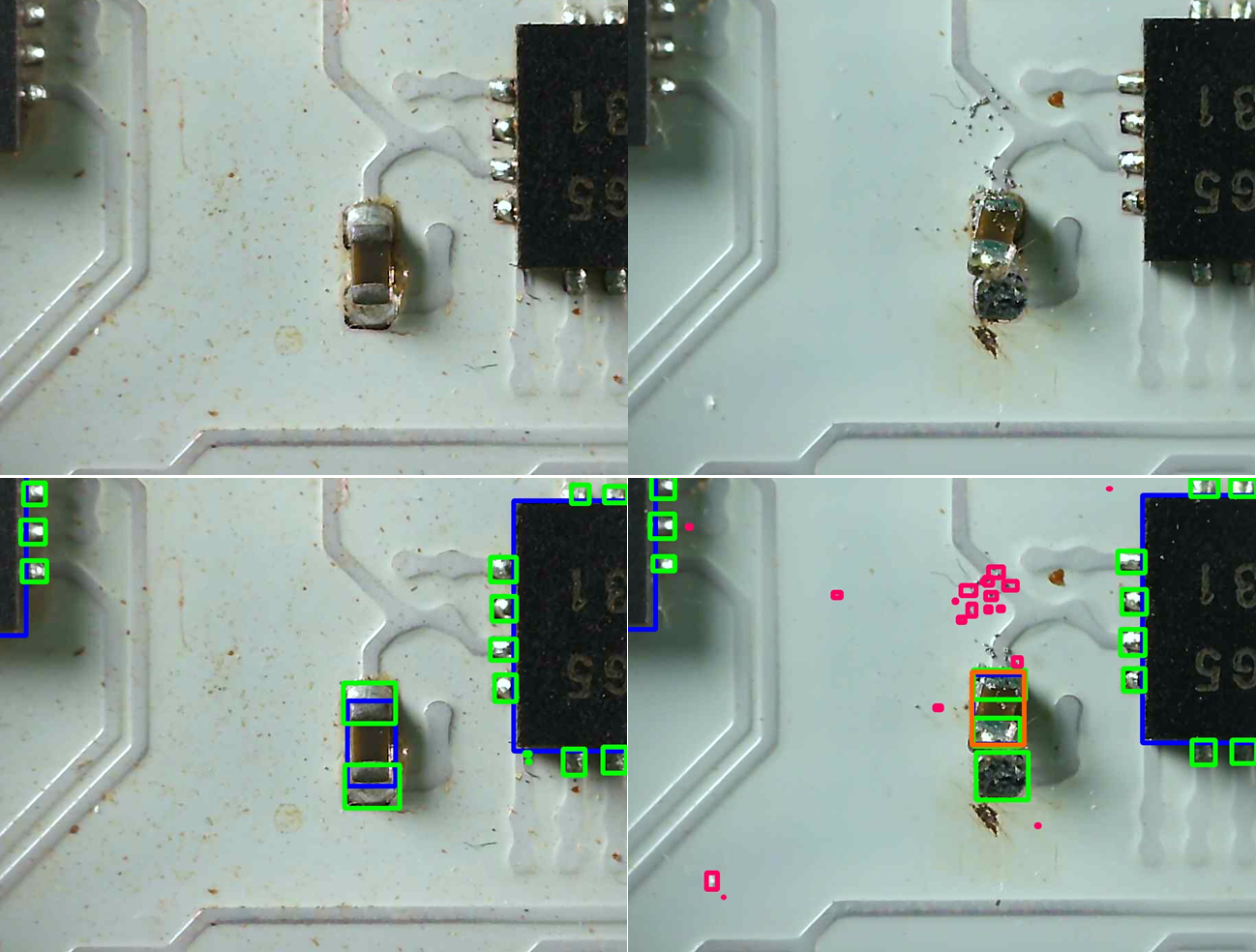}
\caption{Microscope view of an undamaged (left) and damaged (right) satellite board: the trained neural network correctly classifies components (blue) and solderpads (green), as well as the visible defects (tombstone, orange) and pollution (solderball, pink) (lower row)}
\label{fig:aitOptInsp1}
\end{figure}

\begin{figure}
\centering
\includegraphics[width=\textwidth]{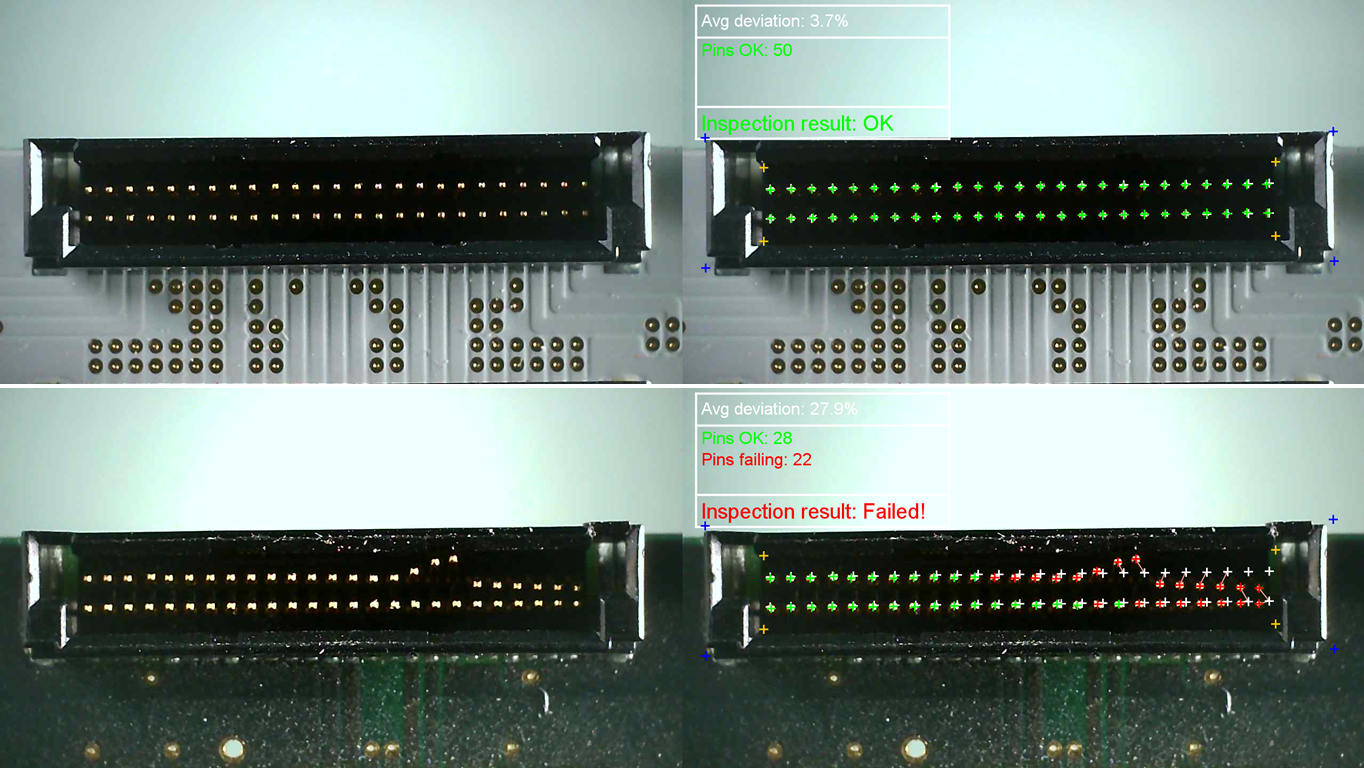}
\caption{Optical inspection of UNISEC connectors: missing, broken or bent connector pins are detected by measuring the deviation of their position from the ideal location relative to the connector casing.}
\label{fig:aitOptInsp2}
\end{figure}

Two examples for the realized AI-supported optical inspection of satellite components are shown in \Cref{fig:aitOptInsp1,fig:aitOptInsp2}. The first image showcases the classification results of surface features on an inspected subsystem PCB board: each subsystem is being inspected before insertion into the satellite bus. The trained neural network determines defects like broken components (so-called \textit{tombstones}) in orange as well as surface pollution in form of solderballs (pink) correctly (\Cref{fig:aitOptInsp1}). If such defects are encountered during the initial optical inspection during the assembly process, this board will then be removed from the production cycle and a spare board be used as a replacement. 

The second image shows optical checks of UNISEC connector plugs on a satellite backplane board that serves as the integration point for subsystem PCBs. Before insertion of any subsystems, each of the connectors on this board is being inspected for missing or broken pins, to avoid damaging the connector of the inserted board as well as non-functional subsystems due to broken connectors. This is done by taking a detailed microscope image of the connector, computing the center of mass of the visible pins, and determining this actual image location with the expected ideal pin position computed by taking into account their locations relative to the also visible connector casing: missing pin centres or large deviations (marked in red in \Cref{fig:aitOptInsp2}) indicate defects of the connector, which would once again lead to the backplane being replaced by a spare board during the production cycle.

This showcases 
how AI functionality can be employed for autonomous inspection of satellite components, determining defects before assembly and thus improving realiability and autonomy of the AIT process. 

\subsection{Learning processes to improve automated robotic integration}
\label{sec:aitLearning}
Main integration step of the developed AIT process is the insertion of the satellite subsystem boards into the backplane panel providing the main assembly structure as well as the common communication bus. The assembly is done by plugging in the UNISEC connector~\cite{unisec2022} present on every subsystem board into the corresponding connector slot on the backplane. This process is automated using the assembly robot. 

The automation solution developed here needs to be highly robust, since in the in-orbit production scenario, no direct human intervention is possible to clear the integration area in case of errors (although teleoperated intervention is a last-resort option). Also, components are not readily available to be replaced. The developed system should thus be able to deal with tolerances and still provide a secure and reliable integration approach.

To ensure this, the robotic assembly is enhanced twofold: first, force supervision is employed to avoid damage of components in case of positioning errors between board and backplane connector. Second, the integration system uses learning methods to learn how to best deal with failed insertions, allowing slight modification of the insertion movement of the robot and such providing a backup-strategy which reliably still allows successful integration even in case of positioning tolerances, making the integration more autonomous and reliable, while also allowing to continuously improve the AIT process using the learning data. This is further detailed in the following sections.

\begin{figure}
\centering
\includegraphics[width=\textwidth]{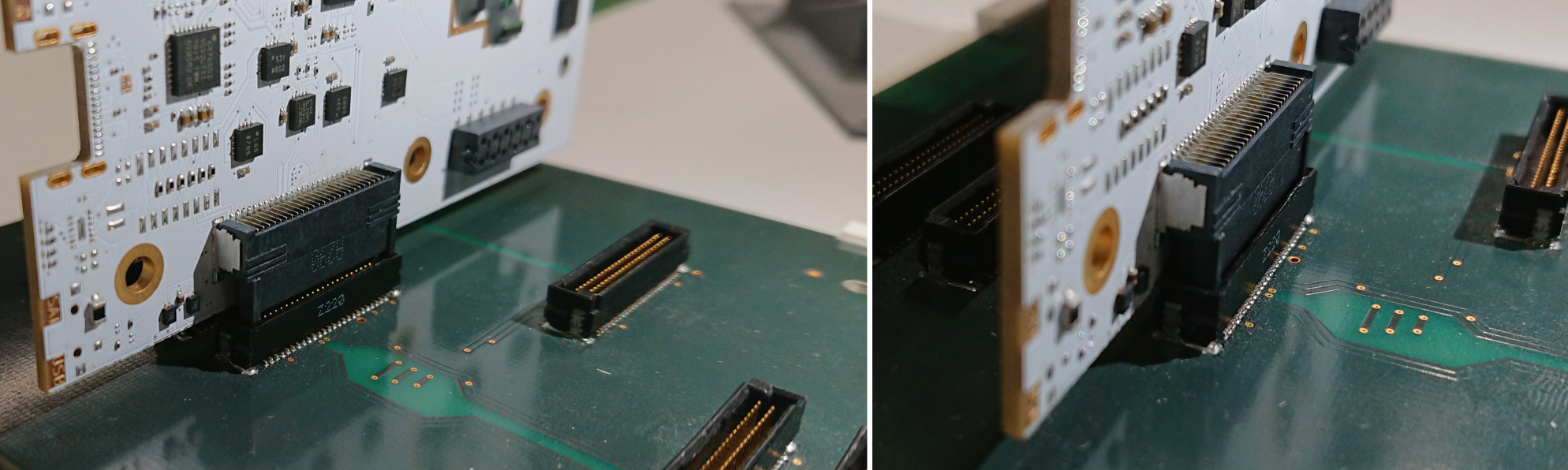}
\caption{Force-sensitive insertion of satellite subsystems: by evaluating forces applied to the board, it can be determined whether connectors are wedged onto each other's casing (left), or freely floating (right) for ideal insertion.}
\label{fig:aitInspPos}
\end{figure}

Safe robotic insertion is realized as follows: the robot first roughly aligns the UNISEC connectors of backplane and subsystem at the insertion position. Then, it slowly drives straight down, permanently evaluating forces applied to the board 
using its internal force-torque-sensors. Should a large force (corresponding to a collision) be encountered during movement it is immediately stopped, and a fall-back insertion procedure started (see the following). At a height where the tip of the subsystem connector should be slightly below the receiving connector casing 
either there are no significant forces acting on the board, meaning that its connector is free-floating and ideally positioned with no connection to the receiving connector's casing; in this case the components are correctly aligned, and insertion can continue. The robot then finishes the insertion by pushing the board further down into final insertion height, in effect gently plugging in the subsystem into the receiving connector on the backplane. However, if at the testing height the forces applied to the subsystem exceed a certain threshold, the connectors are wedged onto each others casing, in which case the robot drives back up to avoid damage, and also starts a fall-back insertion procedure. If a force-spike is encountered during driving to the testing height the connectors might have wedged at first, but then slipped free into proper insertion alignment. In this case, if the force-spike was within tolerable thresholds insertion is continued - otherwise it is also aborted to a fall-back strategy.

Since board and backplane positions are well-known, positioning errors are usually only in the range of sub-millimeters and due to slight bending of the subsystems in the robot's gripper, or slight tolerances of the connector positioning on the board (\Cref{fig:aitInspPos} left shows wedged connectors). The fall-back strategy for insertion is thus to vary the X/Y-position (but not the height above the backplane) of the board relative to its insertion point slightly (usually less than 1mm) and try insertion again at this varied position. In order to not have to blindly try variations, a learning mechanism is employed to determine the most likely successful variation, as well as to continuously improve the insertion process.

\begin{figure}
\centering
\includegraphics[width=\textwidth]{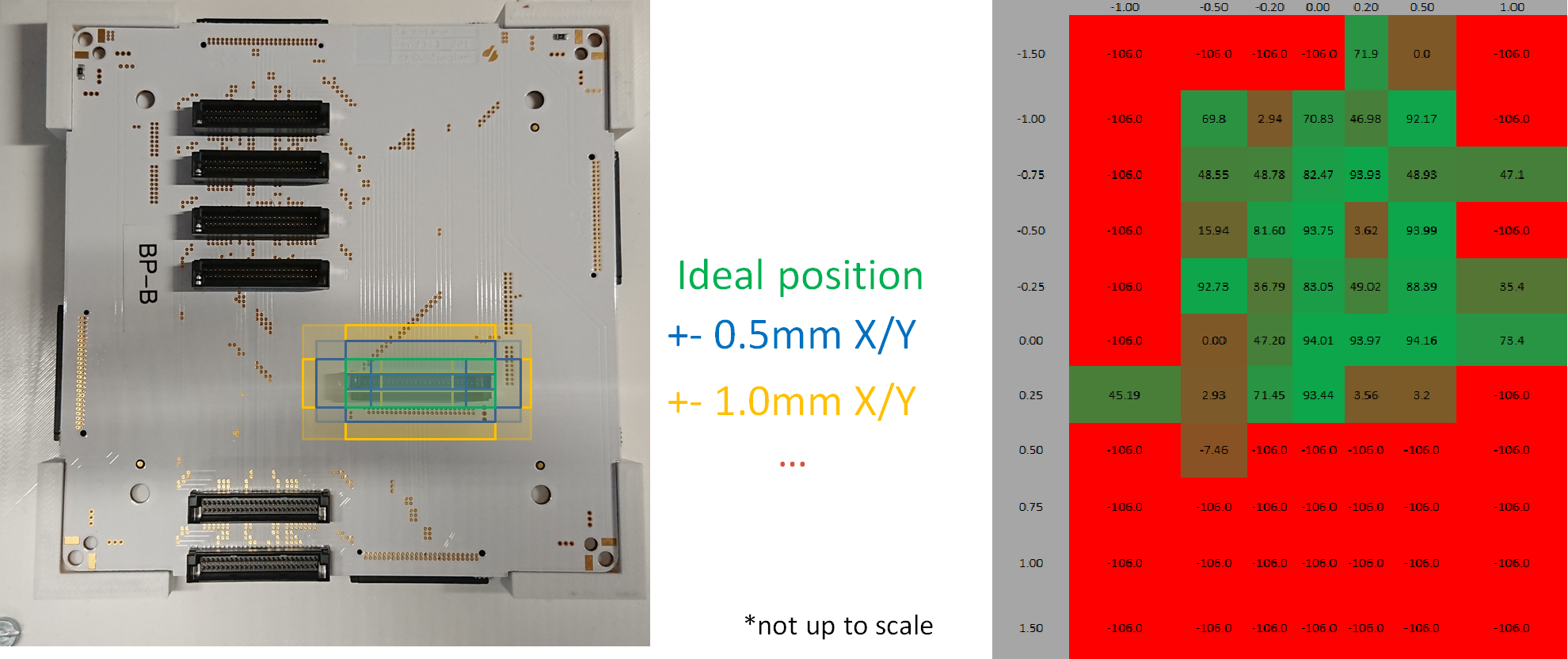}
\caption{Shifting the integration position in incremental steps in X-/Y-direction around the initial, presumably ideal point (left); color-coded learned Q-table for a board type, showing possible insertion positions around the ideal one in the center, and their evaluation score (higher/greener = better, right).}
\label{fig:aitForceExpl}
\end{figure}

For this, a Q-Learning~\cite{dayan1992q,clifton2020q} approach was developed and adopted to the problem at hand: Assumptions are that the optimal insertion position is somewhere centered around the ideal position from setup, insertion height is constant and the board is not rotated relative to the backplane - this is ensured by robot gripper design and backplane mount. Then, a raster of translational shifts in the X- and Y-direction around the ideal position is setup (see \Cref{fig:aitForceExpl}) and stored in form of a two-dimensional table. While inserting the board at one of those shifted positions represented in this table the insertion forces are measured and used to compute an evaluation score: using previous scores as well as measured force values, a value representing how much force was acting on the board at this shifted position is derived. The system was initially trained with \textit{optimistic initial conditions} (meaning each possible shift was tried once first) and then an \textit{epsilon-greedy-strategy} (meaning the system usually chooses the current best value to select the shift position, but with a probability of epsilon chooses a random one to continue exploring and rating different possible insertion positions. By continuously updating the values in the table the robot thus learns which offsets applied usually lead to a successful board insertion (see \Cref{fig:aitForceExpl}).

This force sensitive supervision of the insertion process not only improves safety and reliability by avoiding damage to connectors in case of excessive forces, but coupled with the learning approach also provide an intelligent fallback-strategy in those cases where the initial insertion fails, by providing a look-up option of which shift to the initial insertion position most likely will lead to success (in the experiences made so far). This learning approach was developed using a specific subsystem size and insertion approach, but can later easily be adapted to also handle different types and shapes of subsystems as well as other insertion production steps. Since the modification of the robot's movement is limited to a small area around the initially assumed insertion position, the insertion process still can be executed safely, without risking unexpected large movement of the robot leading to collisions with other subsystems. Continuous updates to the learned Q-learning-table furthermore help optimizing the AIT process, since the initially setup insertion position might not represent a global minimum of the forces applied during insertion - by choosing a different insertion point right away, the forces applied to the satellite subsystems can thus be minimized, leading to less strain on components and thus improving the AIT process. Indeed, with this force sensitive insertion strategy deployed, during project implementation and later test and evaluation runs the robot did not fail for a single time to successfully insert all types of subsystem boards, demonstrating the success of this approach and the thus achieved robustness during automated satellite assembly.

\subsection{Learning systems for electrical fault detection}
\label{sec:aitElectricInsp}
Another check during the automated AIT process was testing of assembled subsystems for proper electronical functionality. Besides testing that integrated boards are powered, connected to the communication bus and functional, this test should also probe and ensure nominal behavior of the board in different system states, by measuring characteristic electric data of the subsystem in those states and comparing them with nominal values provided by the satellite's Digital Twin model. This should allow to detect damaged or worn electronics as well as faulty or missing components on individual subsystems by detecting deviations from nominal power consumption during different subsystem states, and also allow to resolve the (remaining) fitness of the satellite electronics.

To measure electric values from individual subsystems, the AIT process environment included a programmable power supply unit (RIGOL DP832~\cite{rigol2022}) connected to the control PC via a SCPI-interface, allowing to programmably set and readout power values (voltage, current, power) of individual boards, as well as a \textit{DEVBoard} (satellite development board) for communication / control of assembled subsystems.

This control connection with the DEVBoard first allows ensuring principal functionality of the board to begin with, detecting failed integration (UNISEC connector not suitably joined or broken - test by extraction and re-insertion), or damaged/broken boards (disassemble and discard). Furthermore, it allows triggering different functionalities of the subsystems, resulting in different electrical values which can be measured for evaluation. These ``system state'' types depend on the type of subsystem being probed; examples for such states are ``Deactivated'', ``Idle'', ``Computation active'', ``Radio module active'', ``Transmitting'' etc. All system states usually result in different and characteristic current/power consumption levels; these levels were measured and provide the basis for the electrical evaluation.

Although those electrical values are characteristic for a certain system state, they are not necessarily constant and thus cannot be evaluated by simple comparison: they might alternate between multiple values, or slightly fluctuate around certain states; also, while switching between system states, intermittent values will be recorded which must not be used for classification. In order to allow for a proper electronic functionality test and classification, another suitable AI approach was sought-after, allowing to classify readouts into nominal data and outliers indicating abnormal system behavior, possibly due to damaged components or worn electronics. For this classification and outlier detection, several machine learning approaches were evaluated and compared, among others Robust covariance methods and different Support Vector Machines. In the end, the \textit{Local Outlier Factor (LOF)} algorithm~\cite{breunig2000lof} was chosen which uses a density-based approach to detect outliers. The LOF algorithm allows for unsupervised learning of nominal values and is independent of the actual density distribution of the samples, which makes this method best suited to the application at hand. It can first be used to clean the training data of outliers (for example noisy electric measurements between different system states), but also for \textit{novelty detection} - deciding whether a new data set represents an outlier (abnormal system behavior) or nominal values.

The LOF algorithm was first trained with different types of subsystems, toggled between different system states while measuring corresponding electrical data. This data was then preprocessed and stored as nominal values in the Digital Twin for those subsystems. New / different subsystems could then be tested either undergoing the same evaluation procedure and using LOF for novelty detection, but also by computing LOF factors only for individual data points, indicating nominal system behavior or outliers at the read-out values. A suitable threshold for classification was adopted for different system states of the individual boards and allowed to successfully detect abnormal system behavior in simulated electronic defects.

This further electric testing procedure thus showcases another promising application of AI learning approaches to achieve a more dependant and autonomous assembly of satellites. The electronics test could prospectively also be applied to a future in-orbit servicing scenario, testing already used subsystems of satellites in service to determine their remaining fitness.

\section{Teleoperation with adaptive Virtual Fixtures}
\label{sec:teleop}
\begin{figure}
\centering
\includegraphics[width=\textwidth,trim=200px 400px 300px 300px,clip]{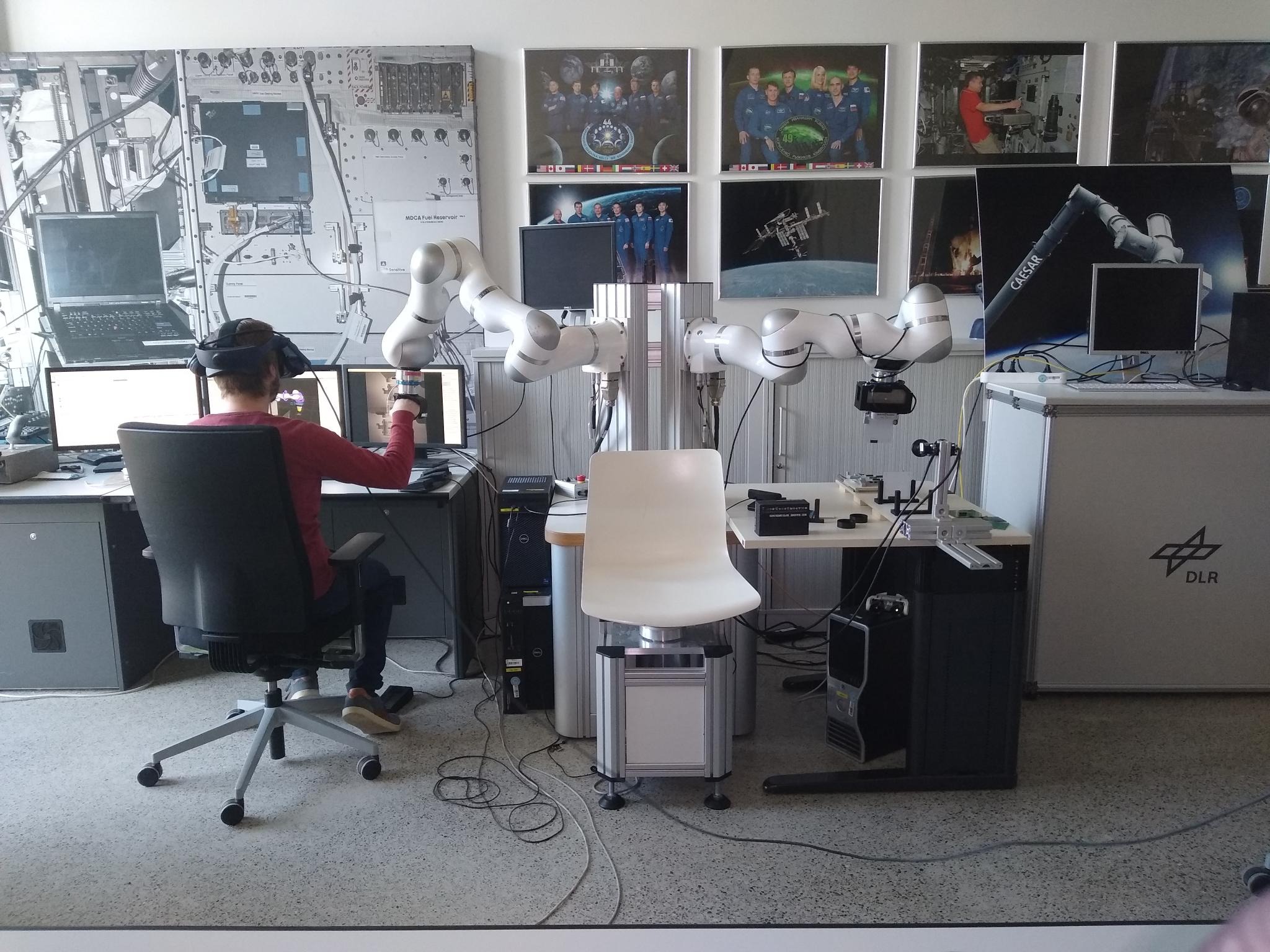}
\caption{The teleoperation system used for CubeSat subsystem assembly in shared control. Identical robot arms are used as haptic input device (left side) and remote robot (right side). A stereo camera setup on the right side allows the human operator to perceive the remote environment while a camera mounted to the gripper enables our vision system to detect target connectors.}
\label{fig:teleop:system}
\end{figure}
In this section, we describe a teleoperation interface based on an adaptive Virtual Fixtures approach \cite{muehlbauer2022multiphase} to facilitate a satellite assembly process that does not require direct human intervention in space, and thus drastically reduces the overall cost~\cite{weber2018space,kempf2021ai}. This interface can be used when the automated assembly and integration fails or to recover from error states. Furthermore, subtasks which have not been automated can be completed using teleoperation. By providing force feedback from both the environment and artificial Virtual Fixtures~\cite{rosenberg1993virtual} along with a video stream, we enable the user to perceive the remote environment and make use of human semantic understanding and dexterity to perform the teleoperation task (\Cref{fig:teleop:system}).

In the following sections, we describe the technical details of our approach (\Cref{sec:mmt}). Its interaction with the Digital Process Twin will be discussed in \Cref{sec:digTwin} and the user interface in \Cref{sec:teleop_interface}. We derive insights which can be used for further improvement (\Cref{sec:teleop_flexibilisation}) from information gained from data recordings (\Cref{sec:teleop_dataset}).

\subsection{Multi-Phase Multi-Modal Haptic Teleoperation}
\label{sec:mmt}
One of the most challenging subtasks in CubeSat manufacturing is the assembly of subsystems requiring a precision of higher than \SI{0.7}{\milli\metre}. As has been previously shown, Virtual Fixtures~\cite{rosenberg1993virtual} providing haptic guidance are essential to successfully complete the task by means of telemanipulation~\cite{weber2018space}.
Virtual Fixtures support the human operator by providing force feedback. They can be classified into ``Forbidden Region'' and ``Guiding'' Virtual Fixtures~\cite{abbott2007haptic, bowyer2013active} where the former prevents users from entering certain regions of the workspace while the latter guides the user along a specific path. Other approaches let the user adaptively define Virtual Fixtures based on camera information~\cite{pruks2022method}, some also explicitly consider time delay~\cite{rosenberg1993use}. As our task is well-specified and we can obtain information about the manipulation target from the Digital Twin, we rely on Guiding Virtual Fixtures in our work.

While Virtual Fixtures based on absolute coordinates can assist the user successfully in many tasks, incorporating additional sensor information is required for highly precise manipulation as in our use case. Camera-based fixture definition \cite{selvaggio2016enhancing} as well as a Visual-Servoing Virtual Fixture \cite{wu2016towards} have already been explored to this end. Both do however suffer from the limitation that fixtures can only be defined when the target is visible in the camera image.

To alleviate this shortcoming and in order to achieve the level of performance required to perform the task of CubeSat subsystem assembly, we use a set of adaptive Virtual Fixtures described in~\cite{muehlbauer2022multiphase} seamlessly combining both a Position-based and a Visual Servoing Virtual Fixture. A globally valid Position-based Virtual Fixture guides the operator to and from the PCB picking position. Once the operator is close to the target and the target connectors come in the field of view, the Visual Servoing Fixture gets gradually activated. Control authority is then linearily blended between both fixtures to ensure a smooth experience for the human operator. Thus, robust and accurate support is provided by the Virtual Fixtures.

For sufficiently accurate positioning of the fixture, information about the location of the PCB to be inserted as well as the expected connector location is queried from the Digital Twin at the start of the teleoperation process. This information is then used to first create a position-based fixture to the subsystem location. After the operator has grasped the subsystem PCB, a new trajectory to the insertion location is generated. During the free motion phase, the Visual Servoing Fixture accurately localizes the grasped PCB to compensate for a possible misalignment. This offset is then added to the fixture being generated when the target connector comes in the field of view. The visual servoing algorithm then continously updates the fixture location as new images arrive. After the PCB has been inserted successfully, the Digital Twin is queried for the next task to perform.

The fixtures are implemented in the real-time controller of the robot which allows target poses to be computed at the control rate of 1kHz. Vision detections arriving at a slower rate of 30Hz are additionally filtered to ensure smoothly varying target poses for the Visual Servoing Fixture. 6-DoF wrenches for impedance control on those target poses are then generated using an angle-axis representation of the orientation~\cite{caccavale1999six,hagmann2021digital}.

For initial experiments with no time delay, a virtual spring-damper system based on the same impedance control scheme as the Virtual Fixtures was used to couple the robots. In case of time delay, passivity-based controllers ensure stability \cite{panzirsch2020safeinteractions,balachandran2020closing}.

\subsection{Augmented Reality Display}
\label{sec:teleop_interface}
\begin{figure}[!h]
\centering
\includegraphics[width=\textwidth,trim=5px 100px 550px 200px,clip]{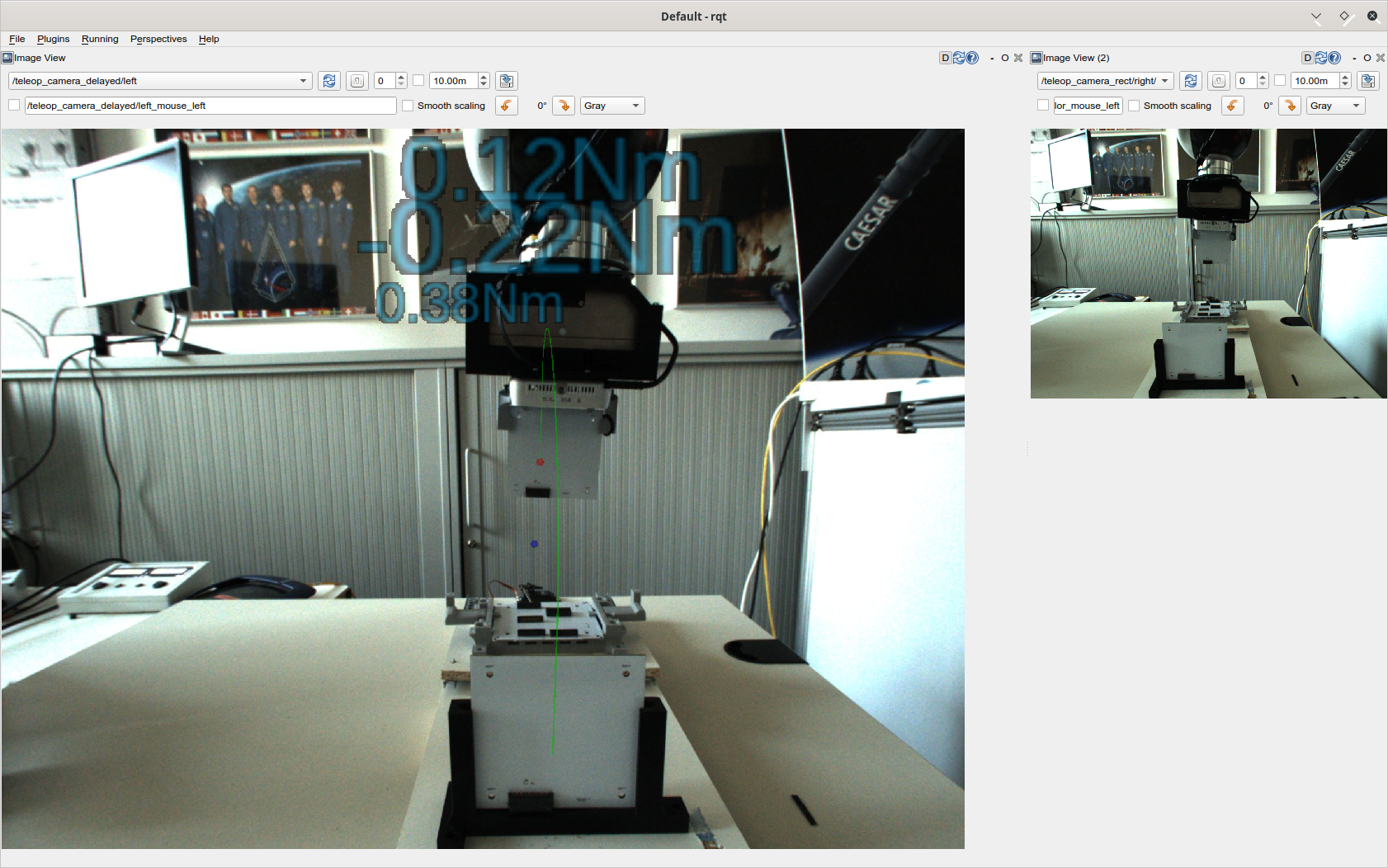}
\caption{Teleoperator view of the assembly scenario through the VR headset shown in \Cref{fig:teleop:system} allows for stereoscopic perception of the remote environment. Rendered overlays of the Virtual Fixtures (green line) and joint torques (blue text) visualize the system state.}
\label{fig:teleop:vrdisplay}
\end{figure}
To improve the operator's perception of the remote environment, we extend~\cite{muehlbauer2022multiphase} by a head-mounted VR display with stereoscopic rendering. Note that by projecting the remote camera view overlaid with Virtual Fixture information, this effectively corresponds to an augmented reality approach where the see-trough capabilities of AR headsets are replaced by a camera stream. \Cref{fig:teleop:vrdisplay} shows the operator's view on the display.

The head-mounted VR display gives a video stream of the remote environment to the human operator. This video stream contains post-processed images taken from a stereo camera pair. One major advantage of using a VR headset in the context of teleoperation is that it allows the human user to perceive the environment in a natural manner similar to the human visual system. Compared to only having several monocular video streams with different views of the environment, providing to each eye an image from slightly different positions allows the human user to have an increased sense of depth perception. Equipped with the ability of binocular vision, the operator should have a better understanding of the remote working environment.

To further guide the human user to perform a successful teleoperation task, the visualization tools available in the simulated environment are leveraged. Instead of displaying raw stereo camera images, a processing step is performed to augment visualizations of the virtual fixtures and vision detections in the video stream projected in the head-mounted display. Therefore, in parallel to the real stereo camera setup, a virtual stereo camera pair with identical camera configurations is also published on RViz\footnote{\url{http://wiki.ros.org/rviz}}. Through the markers visible from the virtual camera pair, the operator is able to receive better visual cues on the state of the system. After both cameras are calibrated and the images rectified, images coming from the real and virtual cameras are overlayed together without the background. The resulting images are then displayed to the human user through the VR headset.

One limitation to note is that the stereo camera pair is fixed to the station with a constant viewing direction of the remote working environment. Therefore, even if the operator moves or tilts their head, their view of the environment does not change which prevents the human user from observing the state of the system from other directions. Further work could thus include mounting the stereo cameras on a pan-tilt unit or to use virtual viewpoints~\cite{lafruit2022immersive}.

\subsection{Teleoperation Dataset}
\label{sec:teleop_dataset}
\begin{figure}
\centering
\includegraphics[width=\textwidth]{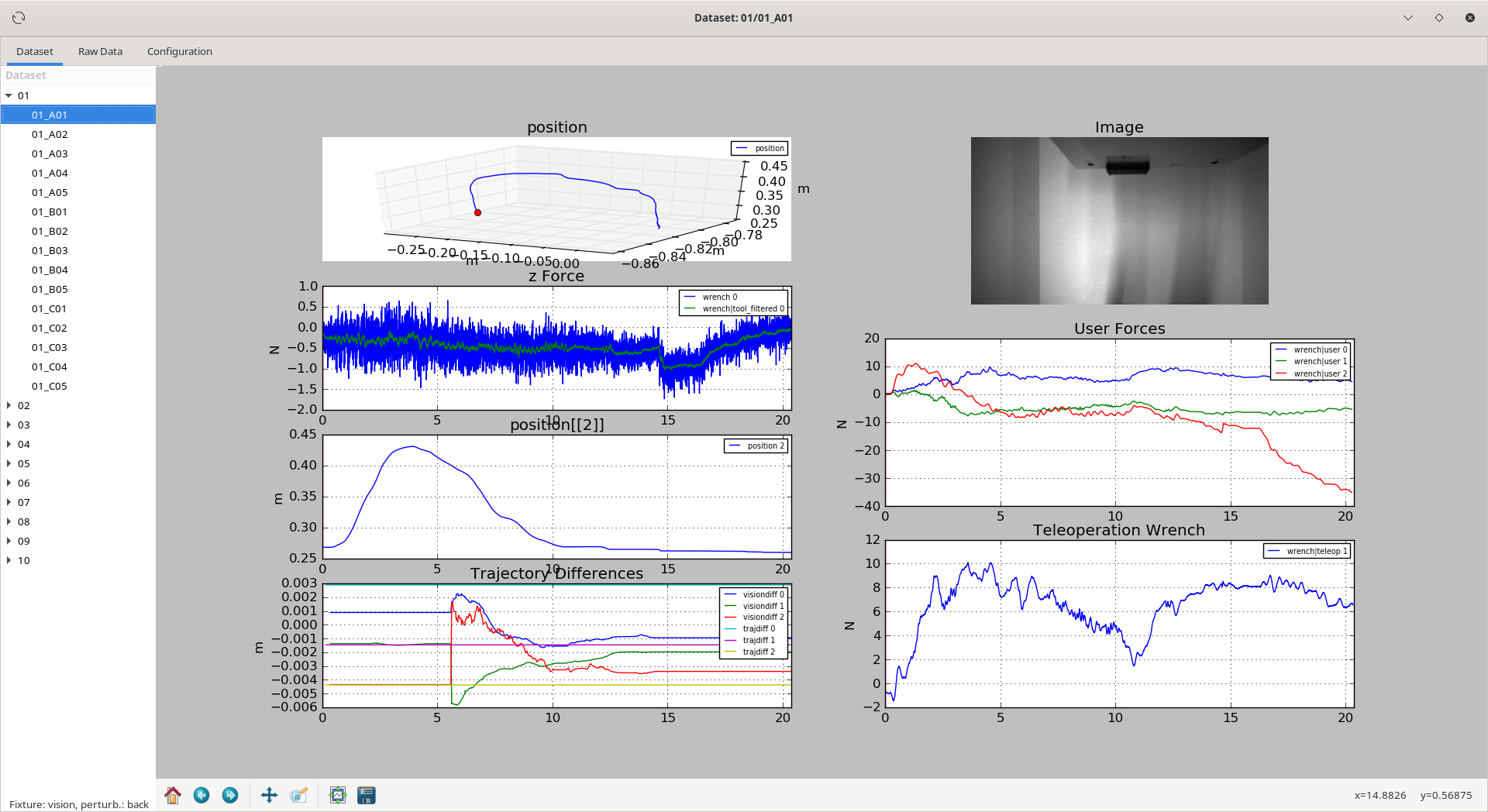}
\caption{Visualization of recorded in our interactive visualization tool. This tool allows to inspect recorded raw data and perform algorithm development (e.g. for the vision system) on these measurements.}
\label{fig:teleop:dataset_tool}
\end{figure}
We use data recorded during the pilot study~\cite{muehlbauer2022multiphase} to enable further algorithmic developments. Recording is performed by logging real-time topic and parameter information from our custom open-source middleware ``links and nodes''\footnote{\url{https://gitlab.com/links_and_nodes/links_and_nodes}}. The parameter information contains hyperparameters used for the visual servoing algorithm as well as stiffnesses used for rendering the Virtual Fixtures. The real-time topic recording contains robot poses, joint torques, input commands of the human operator as well as targets for the fixture. Additionally, images of the in-hand camera as used by the visual servoing algorithm are recorded. \Cref{fig:teleop:dataset_tool} shows an excerpt of such recorded sequence, displaying pose, force and torque and image data.

A flexible class called ``DataContainer'' allows the data to be inspected using different target frames that may also move. It allows both the fixture output generated during teleoperation to be reproduced as well as new approaches to be learned based on user data.

\subsection{Summary and future work}
\label{sec:teleop_flexibilisation}
Our teleoperation approach allows us to place a human at the Earth-based ground station in the loop, should human intervention become necessary during the assembly process. The novelty of our approach is to combine Virtual Fixtures based on different input modalities. This allows us to have the best of both approaches, i.e. the global validity of the position-based VF derived from the digital twin and the high accuracy of the (local) vision-based VF.

The main limitations of this approach are however that we assume an accurate visual perception system and that we use a fixed arbitration function. In future work, we intend to use tools from probability theory \cite{calinon2016tutorial,raiola2018co,zeestraten2017manifold} to take perceptual uncertainty into account to arbitrate between fixtures~\cite{muehlbauer2022mixture} which also require novel passivity-based approaches to ensure stability.

\section{Digital Process Twin with AI-methods}
\label{sec:digTwin}

A Digital Process Twin was developed that represents a virtual image of the AIT process and takes over various tasks in the production of CubeSats. The Digital Process Twin is a special form of a Digital Twin and maps a process virtually and thus has no fixed physical counterpart in the form of a physical twin \cite{kempf2021ai}. In this chapter, the concept of the digital process twin is explained and prototypically implemented. Subsequently, two AI approaches are integrated into the Digital Process Twin and enable on the one hand the customized and automated configuration of the small satellites and on the other hand support the assembly process by an AI-based object recognition method. The developed Digital Process Twin will be verified against the established requirements and will then be validated by means of the demonstrator and the use cases.  

\subsection{Digital Process Twin in production environment}
Following the concepts of Industry 4.0, production systems will evolve into cyber-physical production systems (CPPS) in the future and drive the decentralisation of production \cite{Uhlemann.2017}. These can display the current state of the systems by means of various sensors and data. However, this data can also be used to describe the higher-level processes across systems. By using the data, digital twins are more and more applied in production by enabling advanced production control and optimisation \cite{Kritzinger.2018}. If the actual state data is coupled with digital models of the processes, the Digital Process Twin is created, which not only represents the actual state through simulations, but can also make predictions. Furthermore, all events of Digital Twins in the process are recorded in the Digital Process Twin, which enables detailed analysis. However, the decentralised architecture of CPPS also brings disadvantages. Due to the flexible coupling of the individual systems, classical optimisation methods cannot be used \cite{Ding.2019}. Holistic optimisation in a decentralised production system requires the current process data of the systems involved as well as decision algorithms. These decision algorithms can be implemented with AI methods while aggregating the process data in a digital process twin. The AI methods enable the process twin to constantly learn from the measured data and act predictively in the process.

\subsubsection{Digital Process Twin definition}

\textit{The Digital Process Twin orchestrates the production of a product instance by allocating the relevant product components, instantiating the Digital Product Twin as start of the physical production process and providing information to the Digital Twins of production machines. The structure of the Digital Process Twin is based on the Digital Twin concept with adaptions for process control and functions for a holistic, multi-agent production optimization.}

\subsubsection{Digital Process Twin architecture}
\begin{figure}
\centering
\includegraphics[width=0.87\textwidth]{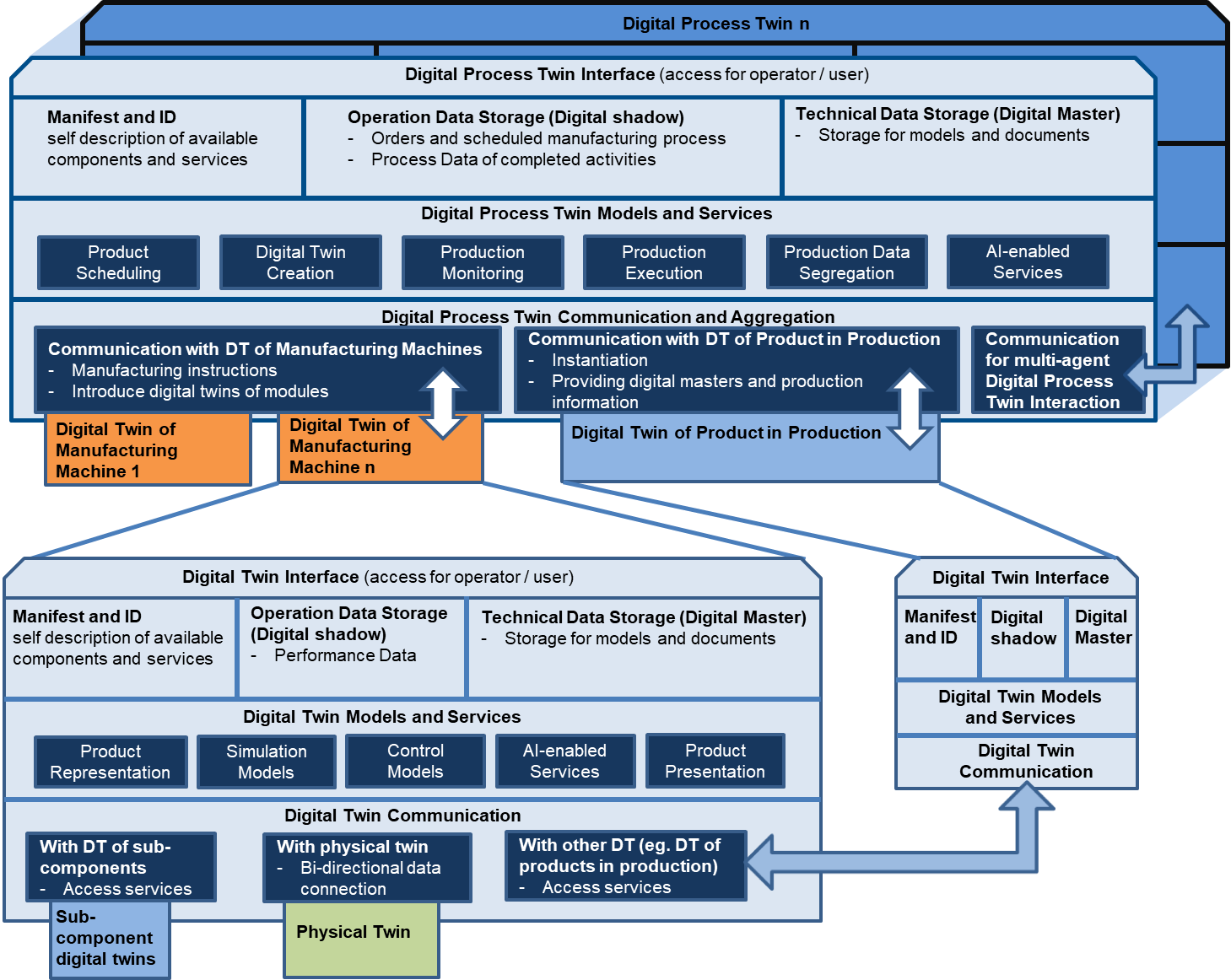}
\caption{Digital Process Twin architecture}
\label{fig:DPTArchitecture}
\end{figure}

Our approach is to combine strengths of hierarchical production analysis with a flexible decentralized production system by transferring the Digital Twin concept into the production process. In this environment, organizational activities and complex decision making require a different set of functionalities compared to product and performance centered functions of Digital Product Twins \cite{Nazarenko2020}. Our Digital Process Twin relies on a Digital Twin architecture to ensure the flexible aggregation and orchestration of Digital Twins of the production system. Inspired by the description of product-service systems of \citet{Tukker2017}, the combination of the process oriented Digital Twin with product oriented Digital Twins, for example of production machines, drives the autonomous production. The architecture of the Digital Process Twin has to reflect the requirements of autonomous production activities coupled with the basic structure of Digital Twins. A schematic overview of the Digital Process Twin and its relation to other Digital Twins is presented in \Cref{fig:DPTArchitecture}.

We divide the Digital Process Twin in the sections interface, stored information, models and services and the communication and aggregation layer \cite{Weyer2016}. The Digital Process Twin interface provides an access point for operator interaction to monitor or control the production process. The manifest and ID section includes information to authenticate and identify the process twin and provide information of available models and services \cite{Boss2019}. The operation data storage (digital shadow \cite{Stark.2020}) is further split into two categories: Data for future activities, such as orders and already scheduled production activities and on the other side data of completed activities. This data is used for evaluation, monitoring and optimization of the production process. Therefore, it differs from the digital shadow of product twins, which have a stronger focus on performance data, for instance to conduct analysis for predictive maintenance of physical components \cite{Stark.2020}. The technical data storage contains the model-based description of the Digital Process Twin, ranging from stored simulation models to trained neural networks.

The execution of models or connected services is situated in the model and service layer. In a Digital Product Twin, relevant models provide the product representation, through CAD models, simulation models such as Simulink or FEM models. In addition, control models for product operation and additional services such as AI-enabled functionalities and product presentation services may be implemented. The Digital Process Twin, however, is equipped with the appropriate models for production planning and scheduling, process analysis and the creation of Digital Twins of products. Other features include the production data segregation, which divides data of the production process for long term archive, process analysis and data that is provided to the Digital Twins as part of their digital shadow.

The Digital Product Twin communication includes the bidirectional interaction with other Digital Twins of manufacturing machines and products. While there is also a real-world production process, the Digital Process Twin only perceives the actual occurrences in the factory through information allocated from Digital Twins with a physical twin. A Digital Product Twin can contain hierarchically structured Digital Twins that map the entire system from different components. The goal of a Digital Process Twin is process control and requires the functionality and information to instantiate Digital Twins of products - an important distinction that is not specified in the current thinking on the use of Digital Twins in a production environment. In combination, the different Digital Twins serve as a foundation for an autonomous production system that can produce new products alongside their Digital Twins. Based on this structure, production flexibility is enhanced by defining a direct interaction between production station and Digital Twin of the product. The two Digital Twins exchange product manufacturing information while the process twin initiates the process, connects component twins and provides services to handle heterogeneous data from different sources \cite{Jasko.2020}.

\subsubsection{Transfer of the Digital Process Twin to the AI-In-Orbit-Factory environment}

AI-In-Orbit-Factory focuses on concepts and solutions for an AI-enabled Cyber-Physical In-Orbit Factory. The autonomous factory in space produces satellites using the standardized modules and interfaces of the CubeSat. This way, fast deployment of satellites is ensured while reducing the required amount of integration tests since modules as opposed to entire satellites are launched into space \cite{WeberMartins.2019}. Since human intervention on site is impossible, the requirements of an in-orbit production pose specifically high demands on reliable and stable processes. At the same time, the nature of mission specific satellite configurations requires an adaptive, product-oriented production process, driving the necessity for product as well as process oriented Digital Twins which is approached using the concept of the Digital Process Twin. While the production is highly automated through adaptive AIT processes, an option for manual intervention is integrated through teleoperation of a robotic arm. In this scenario, the Digital Process Twin provides detailed information of the production task that requires assistance, though some of the information may be conflicting, triggering operator intervention in the first place. Furthermore, relevant information such as performance data of previous production steps similar to the current one is provided. 

\begin{figure}
\centering
\includegraphics[width=0.8\textwidth]{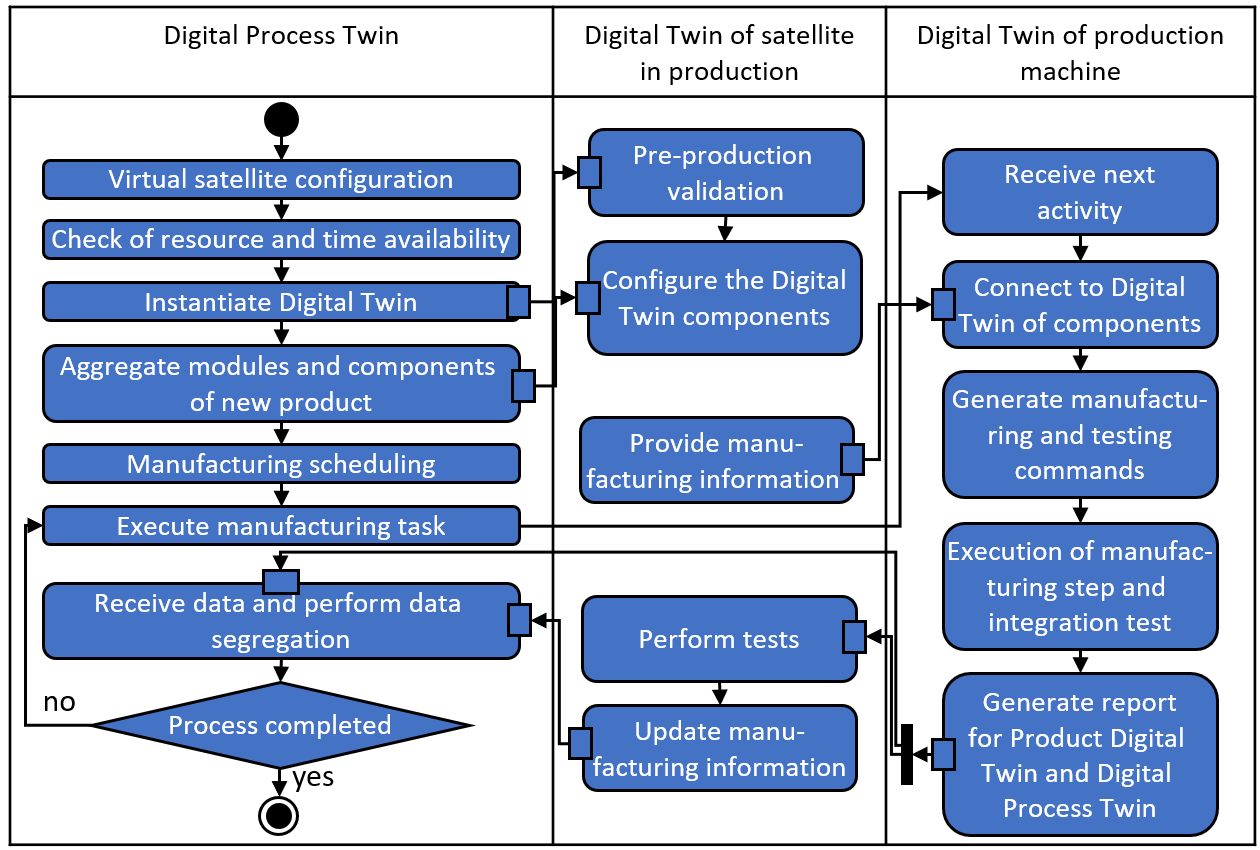}
\caption{UML activity diagram of the in-orbit production process}
\label{fig:DTInteraction}
\end{figure}

Throughout regular production activities, the operation management is distributed, as presented with the UML-activity diagram in \Cref{fig:DTInteraction}. The Digital Process Twin is instantiated from the general process model after receiving the mission requirements and generates a virtual satellite configuration and derives manufacturing information. Afterwards, a check of material and time resources is performed. If the in-orbit factory is able to accept the order, the Digital Process Twin instantiates a generic instance of the Digital Twin of the satellite. The newly instantiated Digital Twin can already perform a pre-production configuration check. Afterwards, the Digital Twin initiates the production process by confirming the valid configuration to the Digital Process Twin. This distinction is significant, as it allows the distribution of pre-production activities and the production control into different systems so that a single unified top-level system is not required by design. Now, the production job is scheduled while the order scheduling remains dynamic to enable a flexible response to changes such as production errors.

Continuous production observation by the Digital Process Twin, Digital Twin of the satellite as well as the production machine ensures high production stability and product quality. The production observation is distributed in three components, accounting for the different kinds of production issues. Here, data of different sources can be used for an informed decision \cite{Schneider.2019}.

\subsection{AI methods for the Digital Process Twin}
The concept of the Digital Process Twin has been extended by AI methods to enable flexibility with simultaneous autonomy. From the multitude of possible AI approaches, two were selected that can be used in the process planning and execution of the Digital Process Twin. Due to the comparatively small amount of data involved in assembling individual CubeSats, many of the approaches that require extensive training data were dropped. To this end, AI-based object recognition was used to monitor the assembly process. Furthermore, the planning of the individual assembly process is automated by using a state machine, which takes over the process planning and execution.

\subsubsection{AI-based image recognition in the assembly process}
\begin{figure}
\centering
\includegraphics[width=0.9\textwidth]{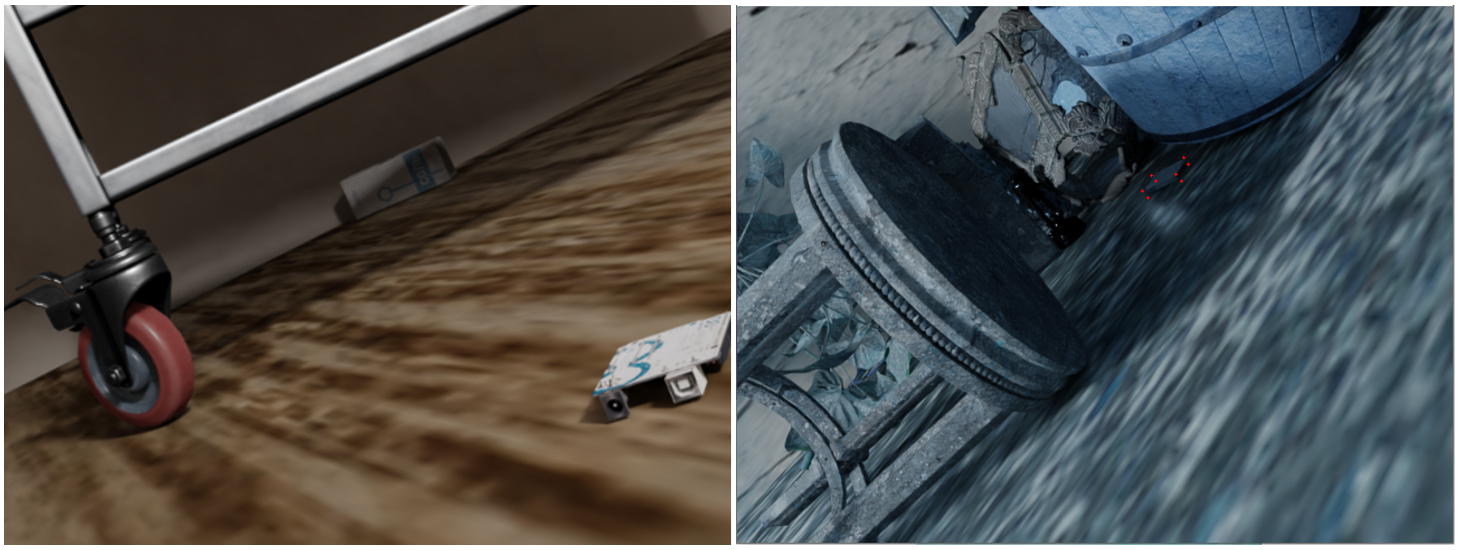}
\caption{Examples of synthetic images for training of the CNN for recognition of electronics}
\label{fig:SyntheticData}
\end{figure}

The AI method used to detected the components of the CubeSats is a Convolutional Neural Network (CNN) as a subcategory of Deep Learning. In this training method, instead of using images of real objects (photographs) from virtual models, generated (synthetic) images are utilized \cite{Hinterstoisser.2019}. The reason for this approach is the lack of training data, which is available for the specific application in the form of the assembly of small satellites in orbit. The process concept must be able to recognize and locate individualized parts of the satellite. The lighting conditions can be neglected, as the in orbit factory offers a closed environment with constant lighting conditions. The goal was therefore the automated generation of application-specific training data. To mitigate the domain gap problem, which occurs between synthetic training data and data from the physical process, domain randomization and photorealistic rendering \cite{Sundermeyer.2020} methods were used. Domain randomization was used in conjunction with 3D rendering \cite{Tremblay.2018}. In photorealistic rendering, a virtual 3D scene is generated and captured by a virtual camera to generate the training images \cite{Hodan.2019}. The \Cref{fig:SyntheticData} shows two examples of synthetic data for training the CNN. For the target objects in the in-orbit factory use case, the CubeSats are electronic components. However, as a first approximation, similar Arduino boards were used instead.

The goal of the object recognition was the identification and determination of the position and orientation in the three-dimensional space of the boards. For this purpose, not only the RGB values of the synthetic image, but also the depth information (depth map) was recorded during the creation of the data sets. Thus, it was possible to determine the 3D position from this 2.5D RGB depth image. Approaches to determine the 3D position using an RGB depth image can be taken from \cite{Deng.2017,Lahoud.2017,Luo.2020,Takahashi.2020}. For localization, a 3D bounding box was placed around the searched object, with the position corresponding to the center of the bounding box. To evaluate the accuracy, the \textit{Intersection over Union} IoU factor was used, which indicates the intersection volume over common (union) volume.

\begin{figure}
\centering
\includegraphics[width=0.9\textwidth]{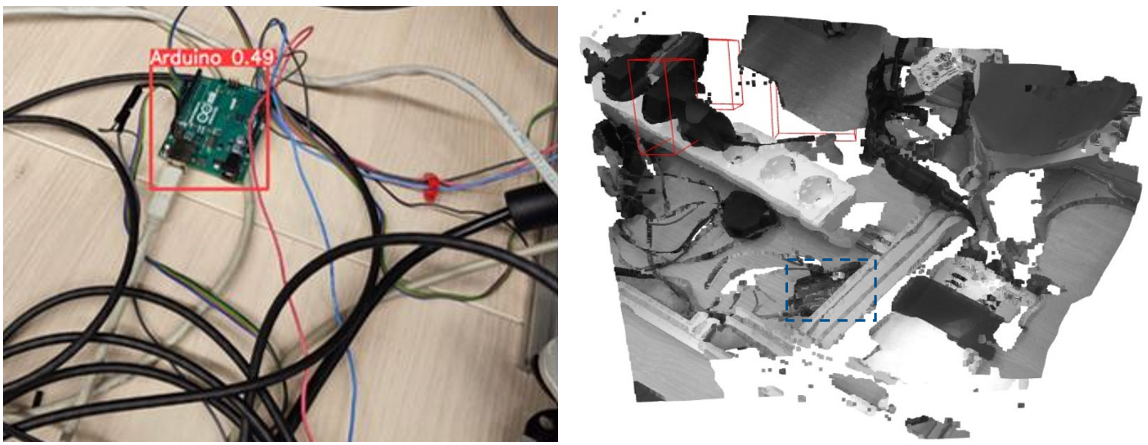}
\caption{RGB picture and depth map of a stereo camera showing an Arduino board}
\label{fig:DepthMap}
\end{figure}

The \Cref{fig:DepthMap} shows an RGB picture and a depth map of a stereo camera. In the depth map, an Arduino is framed as an exemplary electronic component with a blue dashed box. With the implemented AI method, it is possible to monitor the assembly process and provide the Digital Process Twin with additional data from the working environment.

\subsubsection{State Machine for planning and execution of the assembly process}
The assembly process of the CubeSats consists of several assembly steps, which differ in number and sequence depending on the customer's individual configuration. Different boundary conditions have to be taken into account. These include, for example, the assembly space requirements of the individual CubeSat modules, since some modules occupy or cover several slots. But other factors, such as thermal compatibility with regard to dissipated heat, must also be taken into account. 

\begin{figure}
\centering
\includegraphics[width=0.9\textwidth]{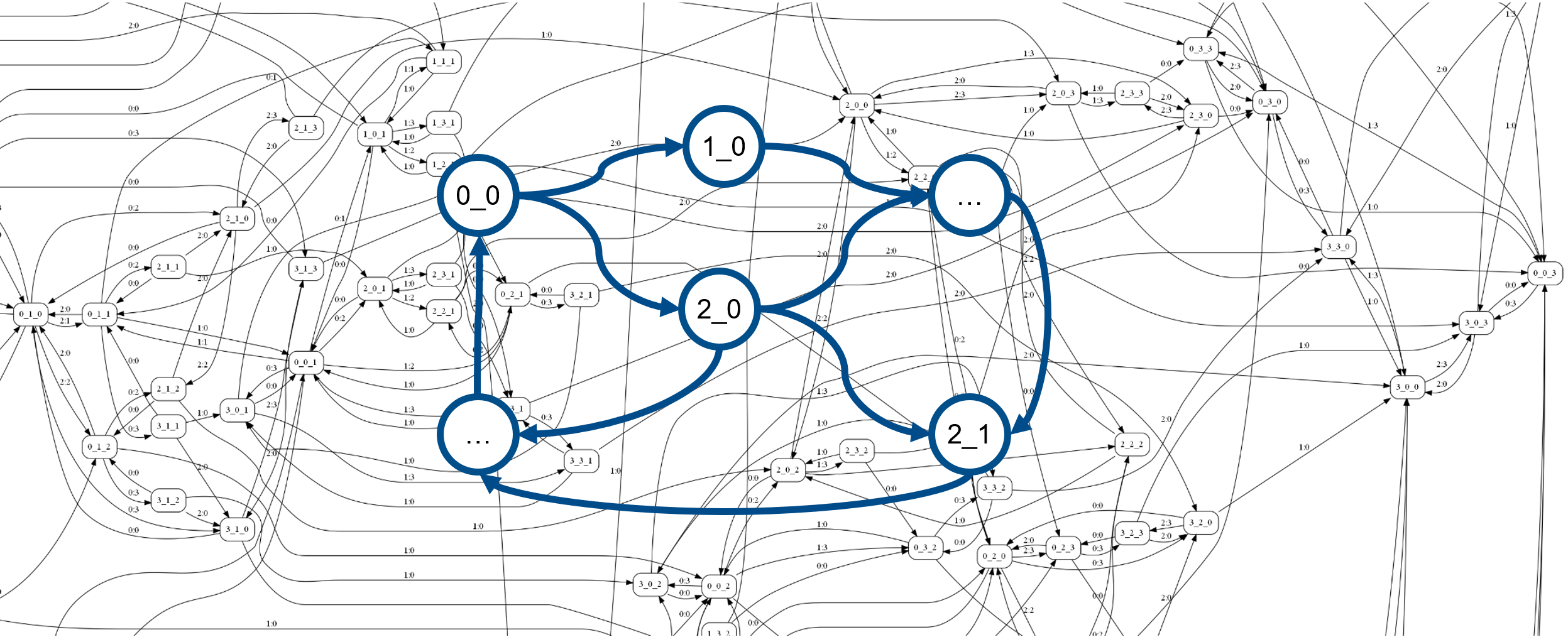}
\caption{State machine for process planing of the assembly}
\label{fig:StateMachine}
\end{figure}

From the configuration of the required functions by the customer and the boundary conditions, a State Machine calculates all possible assembly states and the associated transitions by means of a traversal (breadth-first search). \Cref{fig:StateMachine} shows an exemplary section of a possible state machine graph. Superimposed in the foreground is a simplified example of how it works. The base plate of the CubeSat has several slots for one of the possible CubeSat modules. In the state machine, each state consists of a sequence of numbers, where each digit in the sequence of numbers represents a slot and the number indicates the board mounted on that slot. Using the example of two slots, the number sequence 2-0 would indicate that module number 2 is mounted on the first (front) slot and no module is yet inserted on the second slot. The simplest possible expression of the individual states leads to an optimized speed in the determination of the entire graph, since already two possible slots and two available CubeSat modules result in 3$^{2}$ possibilities. Due to the exponential growth of states and transitions, the computation time increases rapidly. In the background of \Cref{fig:StateMachine}, a graph for 3 slots and 3 CubeSat modules is given.

After generating the entire state machine graph with all possible states and transitions for an individual assembly case, an optimal sequence is determined. The sequence is optimized with regard to the assembly steps. If during assembly the next step cannot be executed as intended, a new path is searched for from this state. An abort can happen, for example, if an error is detected on the board during assembly. The chosen method therefore enables individual, automated, flexible and fault-tolerant planning and execution of the assembly process.

\subsection{Implementation framework of the Digital Process Twin}

\begin{figure}
\centering
\includegraphics[width=0.9\textwidth]{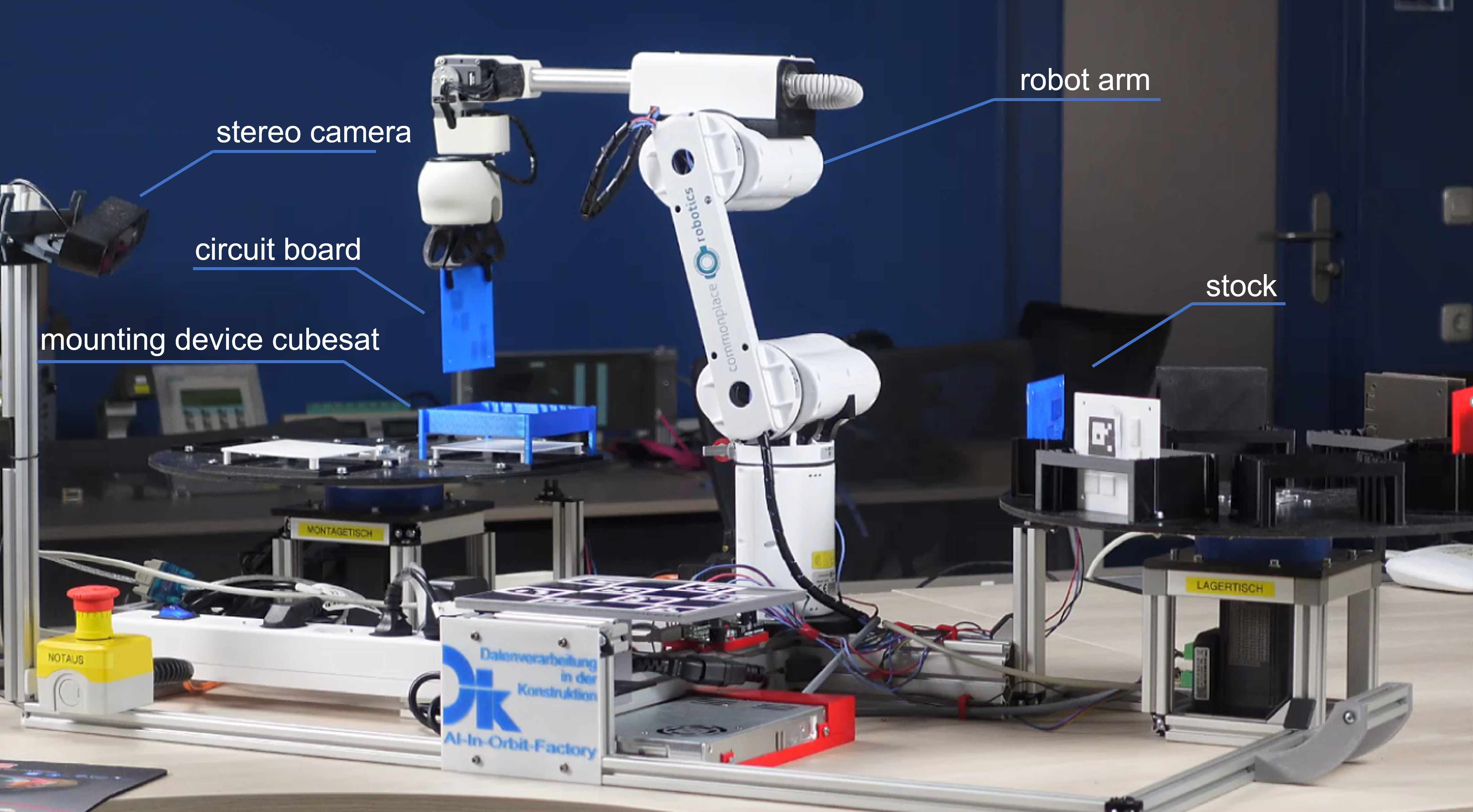}
\caption{Robot arm based demonstrator for the assembly process of the CubeSats with utilization of the Digital Process Twin}
\label{fig:DPTRobotArm}
\end{figure}

The conceptualised Digital Process Twin and the associated AI methods were implemented using the example of a robot arm-based demonstrator \Cref{fig:DPTRobotArm}. A generic software framework of the Digital Process Twin, which is explained below, served as the basis for all functions. The developed framework is designed for flexible use in order to do justice to the versatile architecture of the digital process twin from the \Cref{fig:DPTArchitecture} and the associated diverse functions. The implementation has the five basic functional areas of the Digital Process Twin: \textit{control, transmission, functionality, interface} and \textit{database}.

These basic elements each fulfil specific functions according to which all elements of the Digital Process Twin can be structured. \textit{Control} is a middleware that is used to link the individual functional areas and enables the individual communicate. The tasks of \textit{control} include the mediation between the communication partners, the execution of the individual modules and the communication itself. The mediation serves to identify the individual communication partners and to provide all necessary information (ID, IP address, name, ...). ZeroMQ \cite{hintjens2013zeromq} was used as the communication framework. The data of a process is stored in the \textit{database} area. On the one hand, this includes information about the Digital Process Twin itself and, on the other hand, all data about the process. The \textit{transmission} area is the front-end of the process twin in that it is the only layer visible to the outside world and all interaction takes place via it. Due to the large variety of possible process types, it makes sense to design the functionality of the Digital Process Twin depending on the represented process. For this purpose, they are grouped together generically under \textit{functionality}. 

\subsection{Discussion of the Digital Process Twin}
The developed concept and implementation was verified and validated on the basis of a robotic arm that performs the assembly of a small satellite \Cref{fig:DPTRobotArm}. For the assembly, CubeSat modules are taken from a stock by the robotic arm (Mover6 from Commonplace Robotics), a symbolic test is performed, representing the test procedure from chapter 3, and mounted on a base plate on an assembly table. For the recognition of the real object, a stereo camera (ZED 2 by StereoLabs) was used, which also determines the depth information. In addition to the developed digital process twin framework, other libraries and tools were used as software. In particular, Blender\footnote{\url{www.blender.org}} was used for rendering synthetic data. In the rendering pipeline, BlenderProc \cite{denninger2019blenderproc} do was used to automate the generation of the synthetic data. The generated synthetic data was used to train a PyTorch/YOLOv5\footnote{\url{www.github.com/ultralytics/yolov5}} based AI. The training was done over 300 epochs with synthetic 7000 images and took about 42 hours on a comparatively powerful desktop PC.

To validate in particular the AI approach in the Digital Process Twin using the demonstrator, the AI was trained and used in the assembly process to identify the electronic component and determine its position during handling. As a limit for a correct indication of the determined position, IoU=0.25 was defined in the present case. The object detection reached a Mean Average Precision of approximately 20\% after 300 training epochs. For the process the Digital Process Twin used a state machine to map it. By mapping, AI methods can be used to control the assembly of the individual CubeSats and find the most efficient path. In addition, when errors occur in the process, flexible new possibilities can be found to solve or circumvent the error.

\section{Summary}
\label{sec:summary}
This article presents developments and results of the joint research project ``AI-In-Orbit Factory'', with the focus on realizing an automated production system for small satellites in orbit. It was shown how modern production techniques can be transferred from industry to satellite production, realizing fundamental integration steps of highly modular satellite components using a robotic system as basis. Already during integration, testing steps for the equipment are integrated into the production process. It was then also shown how this AIT process can be essentially enhanced using AI approaches and techniques to improve reliability and autonomy of the production, a necessary improvement for realizing prospective later in-orbit production.

AI-based solutions for optical and electrical inspection of satellite components helped to detect pollution or defects early during production, avoiding producing possibly faulty satellite systems. Learning approaches helped to improve the production process itself, enabling the robot to better handle components during assembly using its force-sensitiveness. For cases where otherwise direct human intervention would be required, a multi-modal shared control approach for teleoperation detailed in \cref{sec:teleop} allows for teleoperated intervention in the assembly and integration process. To recognise when human intervention is required, a digital process twin was created that orchestrates and monitors the individual machines and processes. By means of AI methods, errors can be detected and solutions determined through the virtual mapping of the entire assembly process. 

The system was extensively tested and evaluated during development, showcasing its performance and the usefulness of the developed solution. In future, the proposed system may be improved by incorporating full-fledged FDIR (fault detection, isolation and recovery) techniques to not only detect faulty components, but also autonomously repair and recover from them. Remaining challenge is the qualification of the employed AI methods for usage in space applications. We believe that using our approach of employing AI methods with safety bounds and in non-safety-critical components should significantly simplify qualification.

An extended version of the Virtual Fixtures presented in this work will already be evaluated as part of the SpaceDREAM mission.\footnote{\url{https://www.raumfahrer.net/tag/spacedream/}} An evaluation of the full concept under in-orbit conditions to realize fully automated small satellite assembly in a space factory is left as future work.

\section*{Author Contributions}
FL wrote the abstract, \Cref{sec:introduction}, \Cref{sec:concept}, \Cref{sec:summary} and \Cref{sec:aitProc}. DB, FK and KS contributed to these sections.
MM contributed to \Cref{sec:introduction}, \Cref{sec:concept} and \Cref{sec:summary} and wrote \Cref{sec:teleop}.
VK contributed to \Cref{sec:introduction}, \Cref{sec:concept} and \Cref{sec:summary} and wrote \Cref{sec:digTwin}.
CP contributed to \Cref{sec:introduction}, \Cref{sec:concept}, \Cref{sec:digTwin} and \Cref{sec:summary}.
TD and SD contributed to \Cref{sec:digTwin}.
RA and BS contributed to \Cref{sec:introduction}, \Cref{sec:concept}, \Cref{sec:digTwin} and \Cref{sec:summary}.
BA wrote \Cref{sec:teleop_interface}.
TH contributed to \Cref{sec:introduction}, \Cref{sec:concept} and \Cref{sec:teleop}.
FS contributed to \Cref{sec:introduction}, \Cref{sec:concept} and \Cref{sec:teleop}.
AAS contributed to \Cref{sec:introduction}, \Cref{sec:teleop} and \Cref{sec:summary}.

\section*{Acknowledgement}
The results presented here were achieved within the framework of the AI-In-Orbit-Factory project funded by
the Federal Ministry for Economic Affairs and Climate Action (BMWK) under project numbers 50RP2030A, 50RP203B and 50RP2030C.
We thank João Silvério for feedback on the manuscript.

\appendix



 \bibliographystyle{elsarticle-num-names}
 \bibliography{literature}





\end{document}